\documentclass[twoside,11pt]{article}

\usepackage{jmlr2e}
\usepackage{multicol}
\usepackage{multirow}
\usepackage{array}
\usepackage{hyphenat}
\usepackage{longtable}
\usepackage{graphicx}
\usepackage{parskip}
\usepackage{booktabs} 
\usepackage{amssymb}
\usepackage{amsmath}
\usepackage{subcaption}
\usepackage{float}
\usepackage{adjustbox}
\usepackage{url}

\usepackage{listings}

\usepackage{algorithm, algcompatible}
\usepackage{amsmath}
\DeclareMathOperator*{\argmax}{arg\,max}

\algnewcommand\INPUT{\item[\textbf{Input:}]}%
\algnewcommand\OUTPUT{\item[\textbf{Output:}]}%

\ShortHeadings{A Novel Unfairness Removal Method}{}
\firstpageno{1}

\begin{document}

\title{TowerDebias: A Novel Unfairness Removal Method Based on the
Tower Property}

\author{\name Norman Matloff \email nsmatloff@ucdavis.edu \\
       \addr Department of Computer Science\\
       University of California\\
       Davis, CA 95616 , USA
       \AND
       \name Aditya Mittal \email adimittal@ucdavis.edu \\
       \addr Department of Statistics\\
       University of California\\
       Davis, CA 95616 , USA}

\editor{Unknown}

\maketitle

\begin{abstract}

Decision-making processes have increasingly come to rely on sophisticated machine learning tools, 
raising critical concerns about the fairness of their predictions with respect to sensitive groups. 
The widespread adoption of commercial ``black-box'' models necessitates careful consideration of their 
legal and ethical implications for consumers. When users interact with such black-box models, a key 
challenge arises: how can the influence of sensitive attributes, such as race or gender, be mitigated 
or removed from its predictions? We propose \emph{towerDebias} (tDB), a novel post-processing method 
designed to reduce the influence of sensitive attributes in predictions made by black-box models. Our 
tDB approach leverages the Tower Property from probability theory to improve prediction fairness without 
requiring retraining of the original model. This method is highly versatile, as it requires no prior 
knowledge of the original algorithm's internal structure and is adaptable to a diverse range of applications. 
We present a formal fairness improvement theorem for tDB and showcase its effectiveness in both regression 
and classification tasks using multiple real-world datasets.

\end{abstract}

\begin{keywords}
Algorithmic Fairness, Tower Property, Post-Processing Fairness Techniques, Sensitive Variables, Machine Unlearning.
\end{keywords}

\section{Introduction}

In recent years, the rapid advancement of machine learning algorithms
and their widespread commercial applications has gained considerable
relevance across various fields including cybersecurity, healthcare,
e-commerce, and beyond \citep{Sarker2019intro}. As these models
increasingly drive critical decision-making processes with real-world
impacts on consumers, a noteworthy concern arises: ensuring algorithmic
fairness \citep{Wehner2020intro, Zhisheng2023intro}.  The primary
objective is to minimize the impact of sensitive attributes---such as
race, gender, and age---on an algorithm's predictions. 

A seminal case in the field of fair machine learning is 
the COMPAS (Correctional Offender Management Profiling for
Alternative Sanctions) algorithm by Northpointe, designed to evaluate a
criminal's risk of recidivism. The goal of this tool is to support
judges in making sentencing decisions.

However, the COMPAS algorithm faced scrutiny following an
investigation by \emph{ProPublica}, which presented evidence suggesting
racial bias against Black defendants compared to White defendants with
similar profiles \citep{propublica1}. Northpointe contested these
claims, asserting that their software treated Black and White defendants
equally. In response, \emph{ProPublica} issued a detailed rejoinder,
publicly sharing its statistical methods and findings
\citep{propublica2}. While we do not take a stance on this specific
debate, the case underscores the pressing need to ensure fairness in
machine learning.

A key issue with ``black-box'' algorithms like COMPAS is that users
cannot easily remove sensitive variables $S$ and simply rerun the
algorithm. The source code of the algorithm is typically inaccessible,
and the training data may not be available either. Moreover, there may
be \textit{proxies}, variables that are correlated with $S$, so that the
influence of the latter may continue even though this variable has been
removed.  This raises an important question: how can we mitigate or
eliminate the influence of $S$ in such situations? This paper aims to
address this challenge.

\subsection{The Setting} 

We consider prediction of a target variable $Y$ from a feature vector X
and sensitive attribute(s) $S$. The target $Y$ may be either numeric (in
a regression setting) or dichotomous (in a binary classification setting
with $Y$ = 1 or $Y$ = 0). The $m$-class case can be handled using $m$
dichotomous variables. The sensitive attribute $S$ can be either
continuous, such as age, or categorical/binary, such as gender or race. 

Consider an algorithm developed by a vendor, such as COMPAS, that was
trained on data $(Y_i, X_i, S_i)$, $i = 1, \dots, n$. This data is
assumed to come from a specific data-generating process, with the
algorithm's original goal being to provide the client with an estimate
of $E(Y|X,S)$.  However, in this case, the client wants to exclude $S$
and instead estimate $E(Y|X)$.\footnote{Quantities like $E()$, $P()$,
$Var()$, etc., refer to probability distributions in the data-generating
process, not the sample data.}

In the regression setting, we assume squared prediction error loss to
minimize the error. In the classification setting, we define:

\vspace{-1em} 
\begin{equation*}
  E(Y | X, S) = P(Y = 1 | X, S)
\end{equation*}
\vspace{-1em} 

where the predicted class label is given by:

\vspace{-1em} 
\begin{equation*}
\argmax_{i=0,1} P(Y = i | X,S)
\end{equation*}
\vspace{-1em} 

It can be shown that prediction in this manner
minimizes the overall misclassification rate.

Several potential scenarios for tDB can be identified:

\begin{itemize}

\item [(a)] The clients have no knowledge of the internal structure of the
black-box algorithm and lack access to the original training data. In
this case, the clients will need to gather their own data from new cases
that arise during the tool's initial use.

\item [(b)] Clients have no knowledge of the inner workings of the
black-box algorithm but are given the training data. 

\item [(c)] User of the algorithm, potentially even the original
creator, knows the black-box's details and possesses the training data.
He/she is satisfied with the performance of the algorithm, but desires a
quick, simple way to remove the influence of $S$ in its predictions.

\end{itemize} 

In each setting, the client aims to predict new cases using only $X$.
In other words, although the original algorithm provides estimates of
$E(Y|X,S)$, the goal is to use $E(Y|X)$ as the final predictions
instead. In this paper, we introduce tDB as an innovative approach to
modify the original algorithm's predictions, circumventing the
``black-box'' nature of the model.

\subsection{Paper Outline}

The outline of this paper is as follows. Section \ref{review} reviews the 
existing literature on fair machine learning; Sections \ref{towerDebias_algo}
and \ref{analysis} introduce the \emph{towerDebias} (tDB) algorithm along 
with supporting theory for fairness improvements; Section \ref{empir} presents 
empirical results of tDB across several datasets on both regression and 
classification tasks; Section \ref{discussion} concludes with a discussion 
of various potential societal applications of tDB.

\vspace{-1em}

\section{Related Work}\label{review}

A considerable body of literature has been published in the field of
algorithmic fairness to mitigate biases in data-driven systems across
diverse applications. For example, \cite{chouldechova2017fair} examines
the use of Recidivism Prediction Instruments in legal systems,
highlighting potential disparate impact on racial groups. Collaborative
efforts from the Human Rights Data Analysis Group and Stanford
University have developed frameworks for fair modeling
\citep{lum2016statistical, johndrow2019algorithm}. Significant
research has focused on promoting fairness for binary classification
tasks \citep{barocas-hardt-narayanan, JMLR:v20:18-262}, and 
\cite{JMLR:v25:23-0322} extends these concepts to multi-class
classification. A number of reviews of the existing literature are
available. 

\emph{Machine Unlearning} has emerged as a notable research area focused
on eliminating the influence of undesired factors, such as private data,
copyrighted materials, or toxic content, from a previously trained model
without needing to retrain from scratch \citep{bourtoule2021machine}.
Our tDB method may be view as falling in this research domain. 

Fairness constraints can be integrated across various stages of the
machine learning pipeline \citep{kozodoi2022fairness}. a)
\emph{Pre-processing} involves removing bias from the original dataset
before training the model, with several methods proposed in
\cite{calmon2017optimized, pmlr-v28-zemel13, wisniewski2021fairmodels,
madras2018learning}. b) \emph{In-processing} refers to incorporating
fairness constraints during model training to reduce the predictive
power of sensitive variables \citep{agarwal2019fair}. Notably, a
considerable amount of research in this area involves achieving fairness
through ridge penalties in linear models \citep{scutari2023fairml,
pmlr-v80-komiyama18a, JMLR:v20:18-262, matloff2022novel}. We expand on
this point in Section \ref{komiyama} below. c)
\emph{Post-processing}\footnote{\emph{towerDebias} is post-processing
method.} involves modifying model predictions after training
\citep{hardt2016equality, Silvia2020post}.

In many cases, fairness methods are tailored for a specific machine learning 
algorithm or application. For example, both \cite{zhang2021farf} and 
\cite{zhang2019faht} apply fairness constraints on decision trees; 
\cite{scutari2023fairml, pmlr-v80-komiyama18a} focus on regression models; 
and \cite{chen2024fairness} addresses fairness in Graph Neural Networks for deep 
learning. tDB offers greater flexibility as it can be applied to the predictions 
of any existing ML algorithm and extended to a wide range of applications.

Broadly speaking, fairness criteria can generally be categorized into
two measures: \emph{individual fairness} and \emph{group fairness}. The
central idea behind individual fairness is that {similar individuals
should be treated similarly \citep{dwork2012fairness}, though specific
implementation details may vary. For instance, one approach to achieving
individual fairness is through propensity score matching
\citep{10032333}. In contrast, group fairness requires that predictions
remain consistent across different groups as defined by some sensitive
attribute(s). Common metrics for assessing group fairness include
\emph{demographic parity} and \emph{equality of opportunity}
\citep{hardt2016equality}.

\subsection*{Measuring Biases through Correlations}

Correlation-based measures are frequently used in fairness evaluation
\citep{Deho2022intro, Mary2019FairnessAwareLF}. For instance,
\cite{baharlouei2019r} uses Renyi correlation as a measure of bias to
develop general training frameworks that induce fairness.
\cite{lee2022maximal} introduces a maximal correlation framework for
formulating fairness constraints. Specifically, the Pearson correlation
coefficient offers a way to measure biases between predictions and
sensitive variables in real-world scenarios. \cite{roh2023improving}
explores the effect of changes in Pearson correlation between training
and deployment data in the context of fair training, and
\cite{zhao2022towards} aims to achieve fairness in the absence of $S$ by
minimizing the Pearson correlation between predictions and proxy
variables.

In this paper, we aim to minimize the Pearson correlation coefficient
between our predicted values and corresponding sensitive attribute(s) to
improve fairness by applying tDB. Pearson's correlation is highly
flexible as it allows the evaluation of correlations between dichotomous
and continuous cases of $S$ and/or $Y$ \citep{cohen_applied_2003}. For
categorical or binary $S$, we apply one-hot encoding to create dummy
variables for each sensitive group, then compute the Pearson correlation
to assess individual reductions in correlation. In regression,
$\widehat{Y}$ represents predictions, while in classification,
$\widehat{Y}$ represents the probability $P(Y = 1|X)$.  Lower (absolute
value closer to 0) correlation values indicate reduced association
between predictions and $S$, reflecting improved fairness.

\section{Introducing towerDebias}\label{towerDebias_algo}

In simple terms, \emph{towerDebias} (tDB) estimates $E(Y|X)$ by
modifying the predictions of an algorithm designed to predict from
$E(Y|X,S)$. Our method works as follows:

\begin{quote}
    To remove the impact of the sensitive variable $S$ on predictions of $Y$
    from $X$, average the predictions over $S$. 
\end{quote}  

The Tower Property in probability theory is key here. We will provide a 
formal explanation later, but it essentially states that averaging $E(Y|X,S)$ 
over $S$ while fixing $X$ gives us $E(Y|X)$. Since the latter does not depend 
on $S$, we have effectively removed the influence of $S$. To predict a new 
case, $X_{new}$, we average the predictions $E(Y|X,S)$ across all cases 
where $X = X_{new}$.

Though there are no hyperparameters associated with tDB itself, one
comes in indirectly, as follows.  Computing $E(Y|X)$ using the Tower
Property is an idealization at the population or data-generating process
level, but how do we translate this when working with sample data?

A vendor’s black-box algorithm will estimate $E(Y|X,S)$ using some training 
data. The client may not have access to this data and will need to collect 
their own. Let $(Y_i,X_i,S_i),~ i=1...,,n$ represent the available dataset.

Consider the following example using a dataset that will be referenced 
later in the paper. The dataset includes: $Y$, \textit{a score on the Law 
School Admissions (LSAT) Test}; $X$, a feature vector consisting of 
\textit{family income} (represented in five quintile levels), 
\textit{undergraduate GPA}, and \textit{cluster} (an indicator for the law 
school's ranking); and $S$, \textit{race}. Thus, $X$ = ($income$, $GPA$, 
$cluster$). 

Suppose that we want to predict a new case where $X_{new}$ = ($income$ = 1, 
$GPA$ = 2.7, $cluster$ = 1). Let the race of this individual be African-American. 
Our goal is to generate an $S$-free prediction.

First, we use the black-box model to estimate $E(Y|X,S)$ on the available dataset. 
Using tDB, we average the predicted LSAT scores for all cases in our training data 
that match this exact value of $X$. However, it turns out that there are only 7 such 
cases in our dataset---hardly enough to obtain a reliable average. Furthermore, if 
income were recorded in dollars and cents instead of quintiles, or if GPA were 
reported to two decimal places, we may not have any cases that match exactly this $X$ value.

The solution is to instead average $E(Y|X,S)$ over all cases that are \emph{close} 
to the given value of $X$. To do this, we employ a k-Nearest Neighbor approach, where 
the predictions of $k$ cases with the closest $X$ values to $X_{new}$ are averaged. 
The choice of $k$ affects the bias-variance trade-off and indirectly impacts fairness-utility. 
We explore this point further in the empirical results section.

Section \ref{analysis} provides the theoretical analysis of tDB, including proofs for 
the fairness improvements via correlation reductions. Section \ref{trivariate_deriv} 
quantifies the reduction in Pearson correlation between $E(Y|X)$ and $E(Y|X,S)$ in a 
multivariate normal distribution involving ($X$, $S$, $Y$). Sections \ref{tower_section} - 
\ref{itworksalways} discuss the Tower Property and $L_2$ vector space concepts to show 
fairness improvements achieved with tDB.

\vspace{-1em}

\section{Theoretical Analysis}\label{analysis}

Throughout this section, distributions and random variables are with
respect to the ``population'' or the data-generating mechanism.

Our basic assumption (BA) underlying the steps is that $(Y, X, S)$
follows a multivariate Gaussian distribution, with $Y$ being scalar,
$X$ a vector of length $p$, and $S$ also a scalar. The BA then implies 
that the conditional distribution of any component, given the others, 
is Gaussian, with mean as a linear combination of the conditioning 
components \citep{johnson1993mutli}. 

\subsection{Amount of Correlation Reduction, Trivariate Normal
Case}\label{trivariate_deriv}

Suppose we wish to predict $Y$ using a feature vector containing $X$ and
$S$.  Our goal is to derive a closed-form expression for the reduction
in absolute Pearson correlation between  $E(Y|X)$ and $E(Y|X,S)$. This
reduction will be quantified in terms of the regression coefficients and
the marginal distributions of $X$ and $S$. 

\subsubsection{Problem Setup}

First, let us define key quantities using regression coefficients to
represent the relationships among $X$, $Y$, and $S$.  When using $E(Y|
X, S)$ with a linear model, the predicted $Y$, denoted as
$\widehat{Y_1}$, is:

\begin{equation*}
    \widehat{Y_1} = X\beta + \alpha S
\end{equation*}

where $\beta$ is a vector of length $p$ and $\alpha$ is a scalar. (We
define $X$ to have the constant 1 as its first component, to account for
an intercept term.) Similarly, $E(Y | X)$, denoted as $\widehat{Y_2}$,
can be written as follows:

\begin{equation*}
    \widehat{Y_2} = X\delta
\end{equation*}

where $\delta$ is also a vector of length $p$. Furthermore, the assumption 
of a multivariate normal distribution also allows us to write:

\begin{equation*}
    S =  X\gamma + \epsilon
\end{equation*}

where $\gamma$ is also a vector of length $p$, and $\epsilon \sim 
\mathcal{N}(0, \sigma^2)$ is independent of the predictor $X$.

We aim to derive a closed-form expression for the reduction in
correlation, $\rho_{\text{reduc}}(\widehat Y, S)$,

\begin{equation*}
    \rho_{\text{reduc}}(\widehat{Y}, S) = \rho(\widehat{Y_1}, S) - \rho(\widehat{Y_2}, S)
\end{equation*}

\subsubsection{Compute $\rho_{\text{reduc}}(\widehat{Y}, S)$}
 
\begin{align*}
    \rho(\widehat{Y_1}, S) 
    &= \frac{Cov(\widehat{Y_1}, S)}{\sqrt{Var(\widehat{Y_1})} \sqrt{Var(S)}}  \\
    &= \frac{\beta^T Cov(X, S) + \alpha Var(S)}{\sqrt{Var(\widehat{Y_1})} \sqrt{Var(S)}} \\
    &= \frac{\beta^T  Cov(X) \gamma + \alpha Var(S)}{\sqrt{Var(\widehat{Y_1})} \sqrt{Var(S)}} \\
    &= \frac{\beta^T  Cov(X) \gamma  + \alpha Var(S)}{\sqrt{\beta^T Cov(X) \beta + \alpha^2 Var(S) + 2 \alpha \beta^T  Cov(X) \gamma} \sqrt{Var(S)}} \\
    \\
    \rho(\widehat{Y_2}, S) 
    &= \frac{Cov(\widehat{Y_2}, S)}{\sqrt{Var(\widehat{Y_2})} \sqrt{Var(S)}} \\
    &= \frac{\delta^T Cov(X) \gamma}{\sqrt{Var(\widehat{Y_2})} \sqrt{Var(S)}} \\
    &= \frac{\delta^T Cov(X) \gamma}{\sqrt{\delta^T Cov(X) \delta} \sqrt{Var(S)}}
\end{align*}

Thus, $\rho_{\text{reduc}}(\hat Y, S)$ is:

\begin{equation*}
\rho_{\text{reduc}}(\widehat{Y}, S) = \frac{\beta^T  Cov(X) \gamma  + \alpha Var(S)}{\sqrt{\beta^T Cov(X) \beta + \alpha^2 Var(S) + 2 \alpha \beta^T  Cov(X) \gamma} \sqrt{Var(S)}} - \frac{\delta^T Cov(X) \gamma}{\sqrt{\delta^T Cov(X) \delta} \sqrt{Var(S)}}
\end{equation*}

The formula quantifies the reduction in correlation between $E(Y | X,
S)$ and $E(Y | X)$ in terms of regression coefficients and the marginal
distributions of $X$ and $S$. It demonstrates the potential improvement
in fairness achievable through the application of tDB.

\subsection{$L_2$ Approach}\label{tower_section}

We now formally introduce the Tower Property that underpins our method,
and will be used for that purpose in Section \ref{itworksalways}
These concepts are also relevant in Section \ref{komiyama} and
\ref{itworksalways}, where we show their application in reducing
correlation in a previous paper. 

\subsubsection{Law of Total Expectation and the Tower Property}

The Law of Total Expectation states that if $X$ and $Y$ are random 
variables and $Y$ has a finite first moment, then the following equation
holds:

\begin{equation*}
    E(Y) = E[E(Y | X)] 
    \label{totalexpect}
    \tag{4.1}
\end{equation*}

Note that $E(Y|X)$ is itself is a random variable (a function of $X$), a
key point.

This is a special case of the Tower Property \citep{wolpert2009tower}:

\begin{equation*}
    E(Y | X ) = E[ E(Y| X, S) | X ]
    \label{tower}
    \tag{4.2}
\end{equation*}

and this concept will be central to the methods introduced here.

To illustrate this property intuitively, let’s revisit the LSAT example,
now considering only family income in our $X$. The goal is still to
predict the LSAT score, $Y$, with $S$ representing race. What does
Equation (\ref{tower}) mean at the ``population'' level or the data
generating process?

For example, $E(Y| 4, \text{Black})$ represents the mean LSAT score for
Black test-takers in the fourth income quintile. Now, consider the
following expression:

\begin{equation*}
    E(Y | X = 4 ) = E[ E(Y| 4, S) | X = 4 ]   
    \label{tower4}
\end{equation*}

We first compute the conditional expectation of the LSAT score, $E(Y|X =
4, S)$ for each race and fixed income level $X = 4$ within the inner
expectation. We then average this quantity over $S$, weighted by the
individual probabilities of each race given income level $X = 4$. By the
Tower Property, this weighted average gives us the overall mean LSAT
score for income level 4, regardless of race. 

This aligns with our intuition: conditioning on income (level 4) and
then averaging over all races yields the same result as directly
conditioning on income alone. 

\subsubsection{$L_2$ Operations}
\label{decomp}

As is typical in regression analysis, we assume that all random
variables are centered, meaning their means have been subtracted so that
their expected value is 0. Quantities such as $Y$, $E(Y|X)$, and
$E(Y|X,S)$ are vectors within the $L_2$ space of random variables with a
finite second moment \citep{durrett}, equipped with the 
inner product:

$$
<U,V> ~=~ E(U,V) ~=~ Cov(U,V)
$$

\noindent 
Then also,

$$
||U||^2 = Var(U)
$$

Projections play an important role here; for instance, $E(Y|X)$
represents the projection of $Y$ onto the subspace of mean-zero
functions of $X$. The following property will be key:

\textit{Given random vectors $G$ and $H$ with finite second moments, we
can decompose $H$ as}

$$
H_1 + H_2,
$$ 

\textit{where $H_2 = E(H | G)$ is a function of $G$ and $H_1$ = $H -
H_2$ is uncorrelated with $G$.  Furthermore, $||H_1|| \leq ||G||$.}

These follow from the properties of projections in inner product spaces
\citep{axler}.

\subsection{A Previous Linear Model for Removing the Impact of S}  
\label{komiyama}

Before applying the $L_2$ machinery to tDB in Section
\ref{itworksalways}, let's illustrate the $L_2$ approach in a simpler
context, \cite{pmlr-v80-komiyama18a}. The fairness approach there was
novel, but not rigorous, e.g.\ lacking a statement of assumptions.  Here
we retrace their steps, but in a more precisely stated manner.

For simplicity, assume here that $S$ is scalar, and again that all
variables are centered to mean-0.  Also, continue to assume the BA.

We first regress $X$ on $S$:

\begin{equation*}
    E(X | S) = S \gamma
\end{equation*}


This yields residuals:

\begin{equation*}
    \label{resids}
    U = X - S \gamma
    \tag{4.3}
\end{equation*}


The goal here is to use $U$ instead of $X$ as our feature set. $U$
represents the result of ``removing'' $S$ from $X$. Here's why this
approach works:

%
%

From the results in Section \ref{decomp}, residuals and regressors are
uncorrelated in general (not just in multivariate normal settings). So,
in (\ref{resids}), $U$ is uncorrelated with $S$. In general, zero
correlation does not imply independence; however, under the BA,
uncorrelatedness does imply independence. Thus $U$ and $S$ are
independent, so the use of $U$ as our new predictors will be free of any
influence of $S$, as desired.

%
%
%

\subsection{Proof of General Correlation Reduction, towerDebias}
\label{itworksalways}

With $\rho$ denoting correlation, we wish to show that 

\begin{equation}
\rho(E(Y|X),S) \leq \rho(E(Y|X,S),S) 
\label{wishlist}
\end{equation}

where

$$
\rho(U,V) =
\frac{E(UV)}{\sqrt{Var(U)} \sqrt{Var(V)}}
$$

From Section \ref{decomp}, we know that

$$
\sqrt{Var[E(Y|X))]} = ||E(Y|X)|| \geq ||E(Y|X,S)|| = \sqrt{Var[E(Y|X,S)}]
$$

so that the denominator in the left side of (\ref{wishlist}) is greater
than or equal to that of the right side.

The numerator on the left side is

$$
E[S E(Y|X)] =
E\{E[S E(Y|X) ~|~ S]\} =
E[S E(Y|X,S) ],
$$

and that last term is the numerator on the right side. Thus
(\ref{wishlist}) is established.

\vspace{-1em}
\section{Empirical Study}\label{empir}

This section presents an empirical analysis demonstrating tDB's
effectiveness across five well-known datasets in fair machine learning:
\emph{SVCensus}, \emph{Law School Admissions}, \emph{COMPAS},
\emph{Iranian Churn}, and \emph{Dutch Census}.

\begin{table}[H]
    \centering
    \renewcommand{\arraystretch}{1.5} 
    \setlength{\tabcolsep}{6pt}
    \begin{tabular}{p{4cm} p{3cm} p{3cm} c}
        \hline
        \textbf{Dataset} & \textbf{Response Variable (\emph{Y})} & \textbf{Sensitive Variable(s) (\emph{S})} & \textbf{Rows / Predictors} \\ \hline
        SVCensus & Wage Income & Gender & 20,090 / 5 \\ 
        Law School Admissions & LSAT Score & Race & 20,800 / 10 \\ 
        COMPAS & Recidivism & Race & 5,497 / 14 \\ 
        Iranian Churn & Exited & Gender, Age & 10,000 / 10 \\ 
        Dutch Census & Occupation & Gender & 60,420 / 11 \\ 
        \hline
    \end{tabular}
    \caption{Summary of datasets used in our empirical analysis: response variable, sensitive attributes, and the number of rows and predictors.}
    \label{tab:dataset_table}
\end{table}
\vspace{-1em}

We first train multiple machine learning models to establish baseline
results for fairness and accuracy. tDB is then applied to evaluate
fairness improvements by measuring the reduction in $\rho(\hat{Y},S)$,
while considering potential trade-offs in predictive performance.
Accuracy is evaluated using the Mean Absolute Prediction Error for
regression tasks and overall Misclassification Rate for
classification tasks. Correlation reductions are measured between the
predicted regression values or predicted probabilities $P(Y = 1|X)$ and
the sensitive variable. To minimize sampling variability, we create 25
holdout sets and compute the average test accuracy and correlation
scores.

To establish baseline results, we train several traditional machine
learning models: linear and logistic regression, K-Nearest Neighbors,
XGBoost, Random Forests, and neural networks. The \emph{Quick and Easy 
Machine Learning} (\texttt{qeML}) package in R provides a user-friendly 
framework for generating predictions with linear and logistic regression, 
K-Nearest Neighbors, XGBoost, and Random Forest algorithms. The neural 
network is trained using the \emph{PyTorch} package in Python.

In addition, we compare our method by applying it to the fair ridge
regression (FRRM) and fair generalized ridge regression (FGRRM)
algorithms presented in \cite{scutari2023fairml}. The fact that tDB is a
post-processing method suggests that the fairness achieved by FRRM/FGRRM
might be further improved by using \textit{both} methods. In other
words, one might perform our initial analysis using, say, FRRM,
producing a rather fair result, and then achieving further improvement
by applying tDB.

Both FRRM and FGRRM enforce fairness using \emph{Equality of
Opportunity} as the chosen criterion and result in lower initial
correlations between predictions and the sensitive variables.  These
algorithms incorporate an unfairness parameter to control the degree of
fairness, and we evaluate our method across several different
levels.\footnote{To be sure, users can achieve a similar
fairness-utility trade-off by adjusting the unfairness parameter
directly. Our goal is to evaluate how tDB performs at a given unfairness
level, evaluating it across multiple values.} This approach integrates
fairness directly into the training process, allowing us to assess
whether tDB further reduces correlations in the post-processing stage.

The results of our analysis are presented graphically, highlighting
\emph{comparative} trade-offs between accuracy losses and correlation
reductions. \textbf{Note:} The graphs display relative changes in
fairness and accuracy compared to the baseline models. For example, if
k-Nearest Neighbors has an initial misclassification rate of 0.2, the
left vertical axis reflects that value, not 0. The same applies to
correlation reductions. Our goal is to show the improvements in fairness
(and corresponding accuracy trade-offs) achieved by tDB over existing
methods. 

A key question in our experiments is selecting an appropriate value for
$k$. Our method does not have an inherent ``slider" for adjusting the
Fairness-Utility Tradeoff.  However, the k-Nearest Neighbor approach
involves a Bias-Variance trade-off as we average the estimated
$E(Y|X,S)$ over $k$ nearest data points given value of $X$. A small $k$
results in higher variance, while a large $k$ might include data points
far from $X$ and induce bias.  This may cause an indirect
Fairness-Utility trade-off, as follows.

Specifically, a large value of $k$ has an effect rather similar to an
increase in the variance of $X$. Consider for example a setting in which
$X$ and $S$ are independent. As we increase $Var(X)$, the role of $S$ in
$E(Y|X,S)$ declines, thereby lowering $\rho(\widehat{Y},S)$.  However,
this comes at the cost of predictive accuracy, i.e., utility. Hence,
what may seem like a Fairness-Utility trade-off is actually the
Bias-Variance trade-off inherent in k-NN. The experiments provide
insights into this trade-off to help determine the optimal choice of $k$ when applying tDB.


\subsection{SVCensus}

The \emph{SVCensus} dataset, a subset of early 2000s U.S. Census data,
focuses on income levels across six engineering occupations in Silicon
Valley. Each record includes attributes like occupation, education,
weeks worked, and age. Our goal is to predict wage income ($Y$) in a
regression task, with gender ($S$) as the sensitive attribute. 

Figure (\ref{fig:svc_ml}) shows the impact of tDB on both fairness improvements and 
corresponding accuracy losses across various values of \( k \). In terms of accuracy, 
the baseline MAPE for each algorithm is approximately \$25,000, with minimal additional 
accuracy losses after applying tDB. Even at higher values of \( k \) (\( k \geq 30 \)), 
the accuracy loss remains below 5\% (less than a \$1000 increase) compared to the original
predictions. Regarding fairness, the baseline correlation between predicted income and 
gender for each model is around 0.25, with tDB reducing it by up to 50\%, bringing it 
below 0.1 starting at \( k \geq 10 \). Our results show that tDB substantially reduces 
gender-income correlations while maintaining minimal accuracy losses.

Interestingly, with XGBoost, accuracy improves by \$1,000 after applying tDB. Combined 
with a correlation reduction of up to 50\%, this shows a significant improvement in 
both fairness and accuracy. 

\vspace{0.5em}
\begin{figure}[H]
    \centering
    \includegraphics[width=1\linewidth]{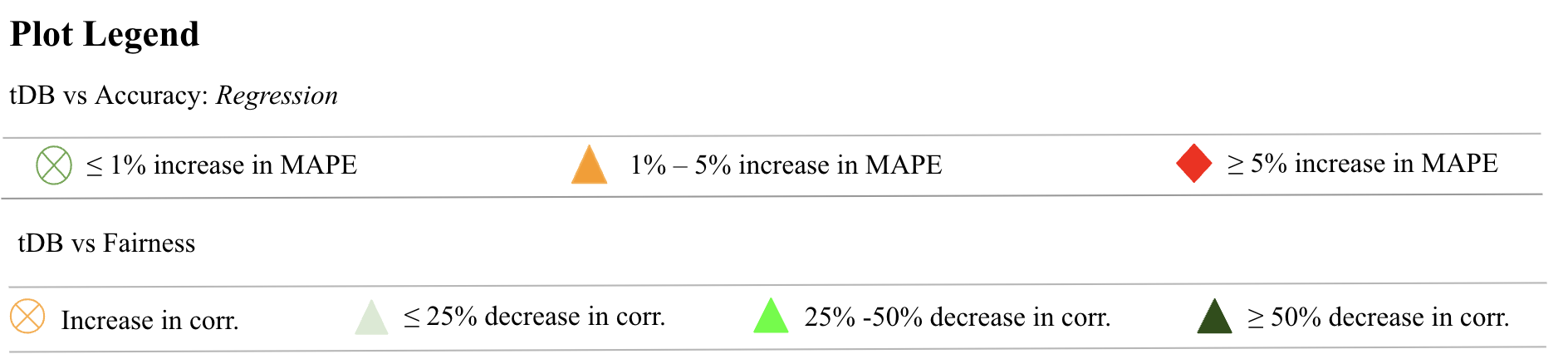}
    
    \caption*{ML versus \emph{towerDebias}: SVCensus Results}
    \vspace{0.5em}
    \includegraphics[width=1\linewidth]{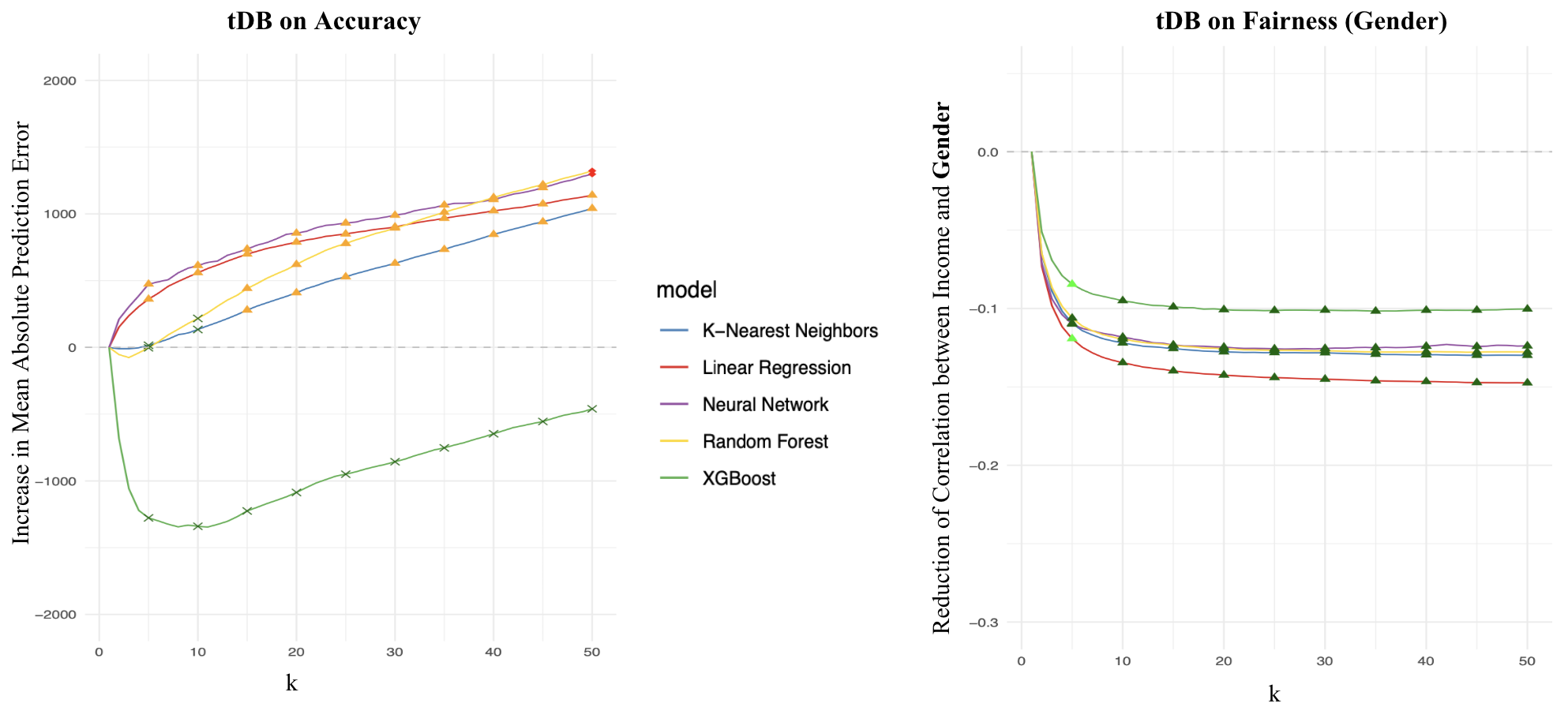}
    \vspace{-0.5em}
    \caption{Effect of \emph{towerDebias} on MAPE increase on predicting \textbf{wage income} and the correlation reductions with \textbf{gender} in the SVCensus dataset.}
    \label{fig:svc_ml} 
\end{figure}

\subsection{Law School Admissions}

The Law School Admissions dataset (1991) surveys U.S. law students, capturing demographics 
and academics like age, GPA, income, gender, and race for admitted applicants. The goal is 
to predict LSAT scores ($Y$) in a regression task, with race ($S$)—categorized as Asian, 
Black, White, Hispanic, and Other—as the sensitive attribute. We now extend fairness analysis
beyond binary cases of $S$.

\newpage
\begin{figure}[H]
    \centering
    \caption*{ML versus \emph{towerDebias}: Law School Admission Results}
    \vspace{0.5em}
    \includegraphics[width=0.95\linewidth]{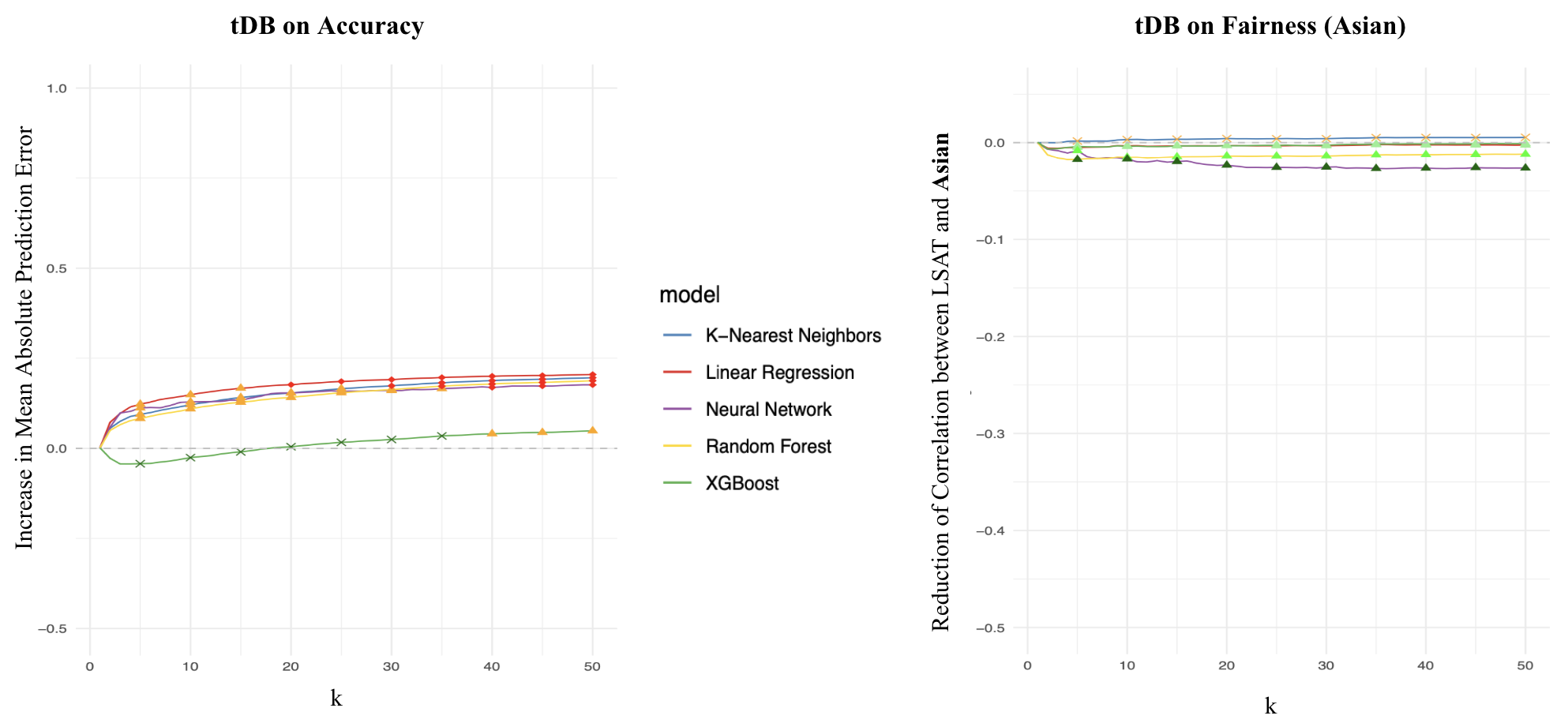}
    \vspace{-0.5em}
    \includegraphics[width=.95\linewidth]{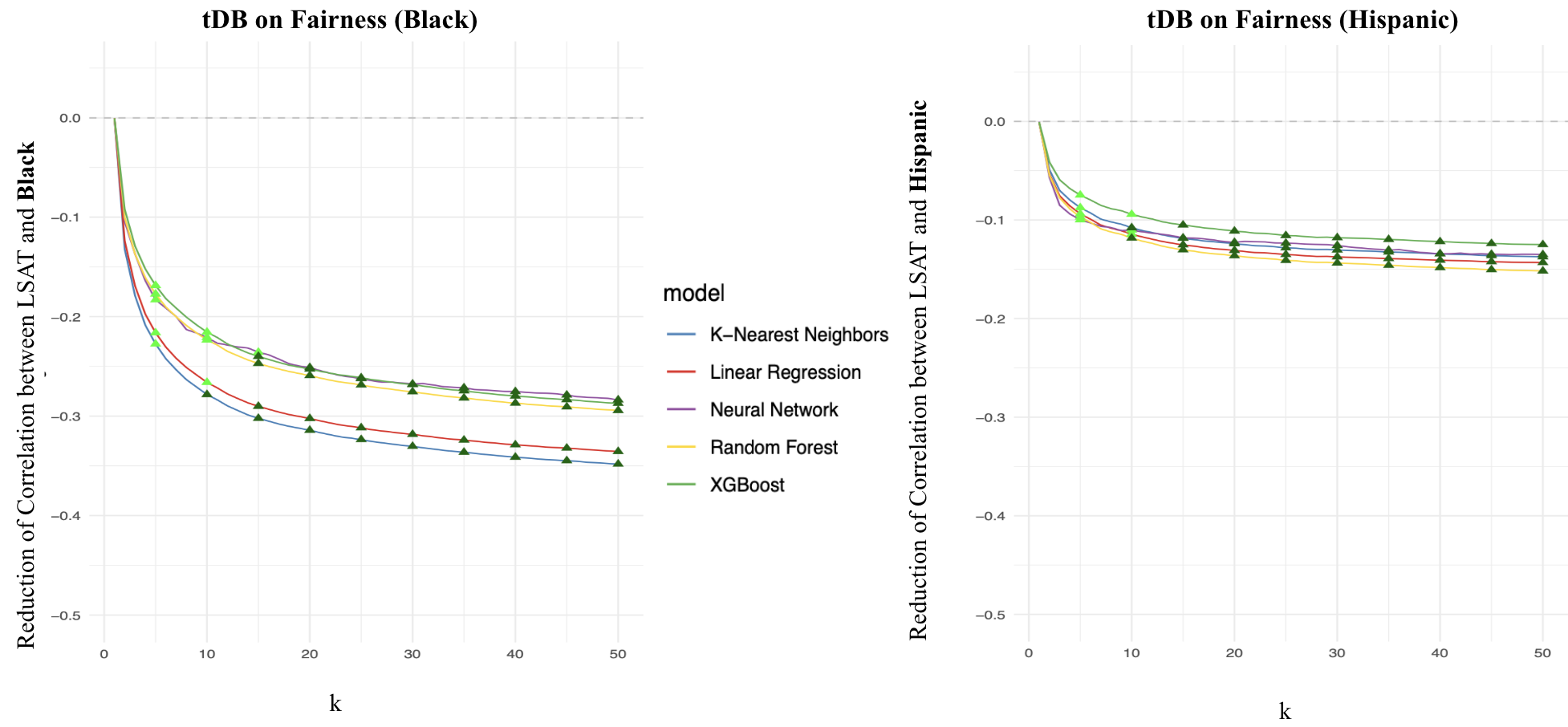}
    \vspace{-0.5em}
    \includegraphics[width=0.95\linewidth]{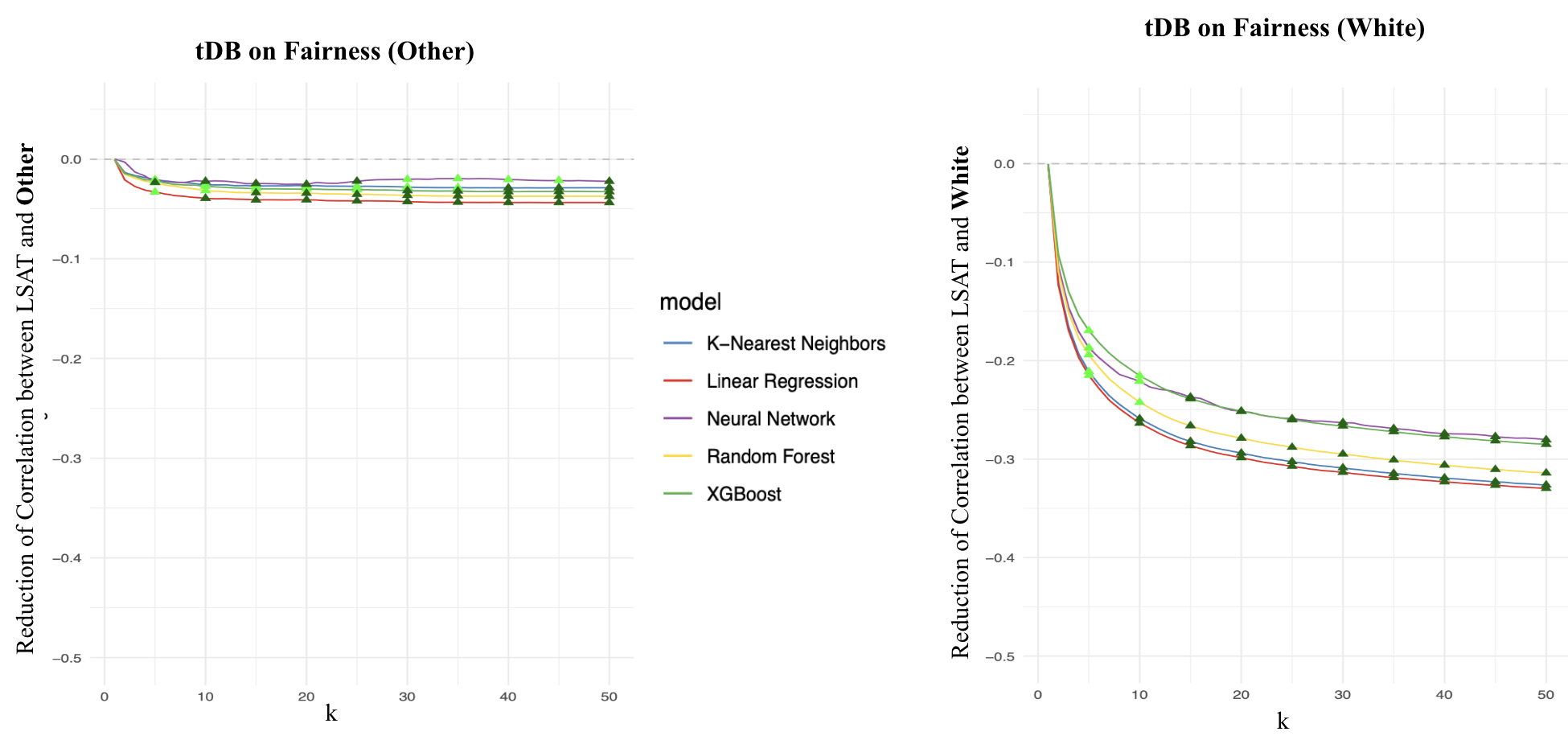}
    \vspace{0.5em}
    \caption{Effect of \emph{towerDebias} on MAPE increase on predicting \textbf{LSAT Score} and correlation reduction with \textbf{race} in the LSA dataset.}
    \label{fig:lsa_ml} 
\end{figure}
\newpage

Accuracy results in Figure (\ref{fig:lsa_ml}) are similar to those in
the \emph{SVCensus} dataset. Baseline algorithms show an average MAPE of about 3.5 
points, with tDB causing a 5\% additional loss in accuracy (less than 0.5 point 
increase in MAPE) across moderate-higher values of $k$. In terms of fairness, each 
racial group shows different baseline correlations with predicted LSAT scores, with 
tDB achieving substantial reductions. For instance, in the neural network model, the 
correlation between predicted LSAT scores for the ``African-American'' and ``White'' 
groups  was approximately 0.3, for the ``Asian'' and ``Other'' groups it was below 0.1, 
and for the ``Hispanic'' group it was 0.15. Similar patterns can be observed from the 
other baseline machine learning models. Across all methods, applying tDB resulted in a 
reduction of more than 50\% in the correlations in all racial groups from early to 
moderate values of the trade-off parameter $k$.




\subsection{COMPAS}

The \emph{COMPAS} dataset contains data on criminal offenders screened
in Florida (2013-14) and is used to predict recidivism, denoted as $Y$,
indicating whether an individual will recommit a crime in the near
future. Race is the sensitive variable ($S$), with three categories:
White, African-American, and Hispanic. The dataset includes features
such as criminal decile score, offense date, age, number of prior
offenses, sex, and more. The task is now a binary classification problem
with race as a categorical sensitive variable.

\begin{figure}[H]
    \centering
    \includegraphics[width=1\linewidth]{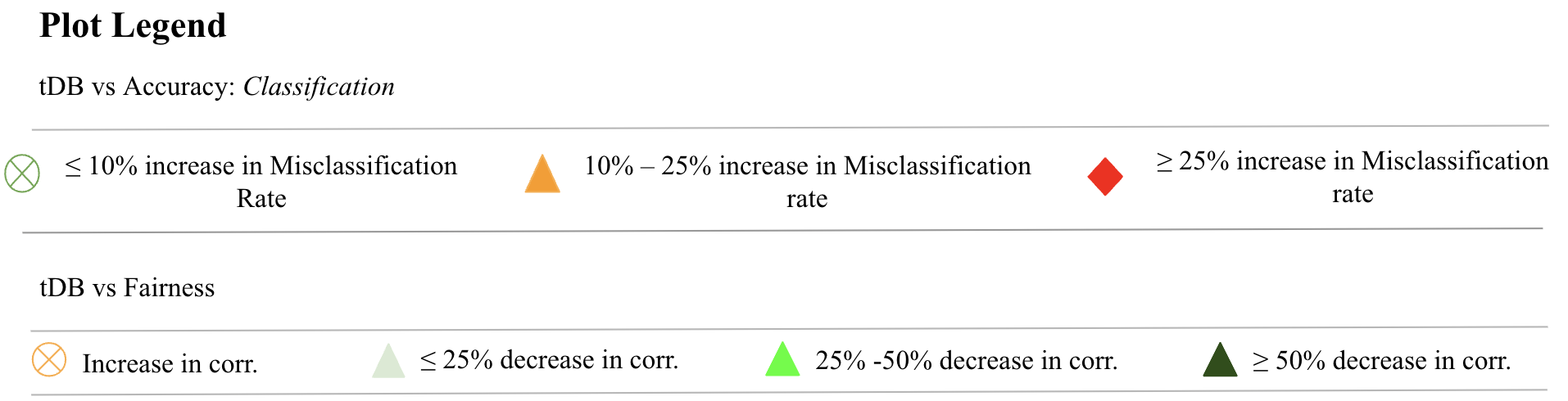}
    \caption*{ML versus \emph{towerDebias}: COMPAS Results}
    \includegraphics[width=0.95\linewidth]{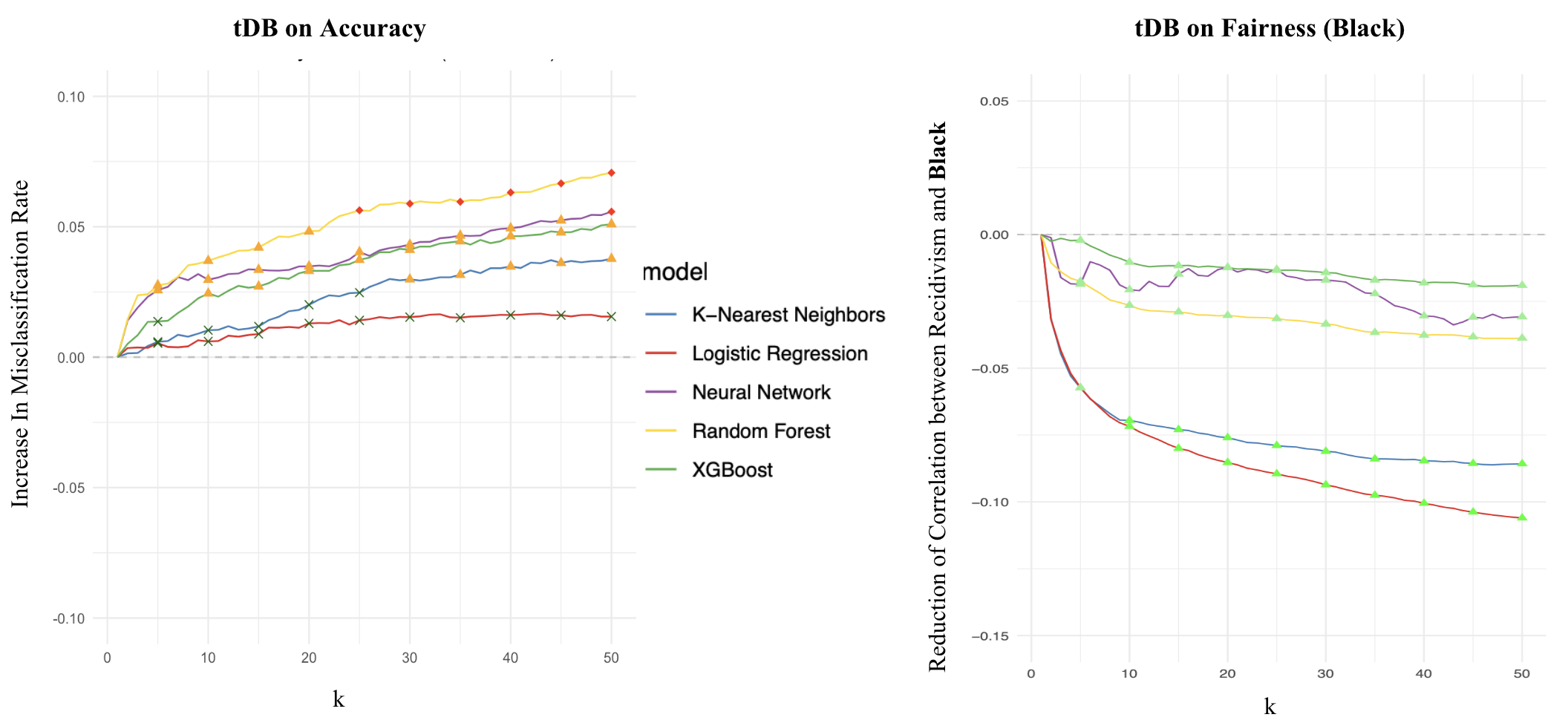}
\end{figure}

\begin{figure}[H]
    \centering
    \includegraphics[width=.95\linewidth]{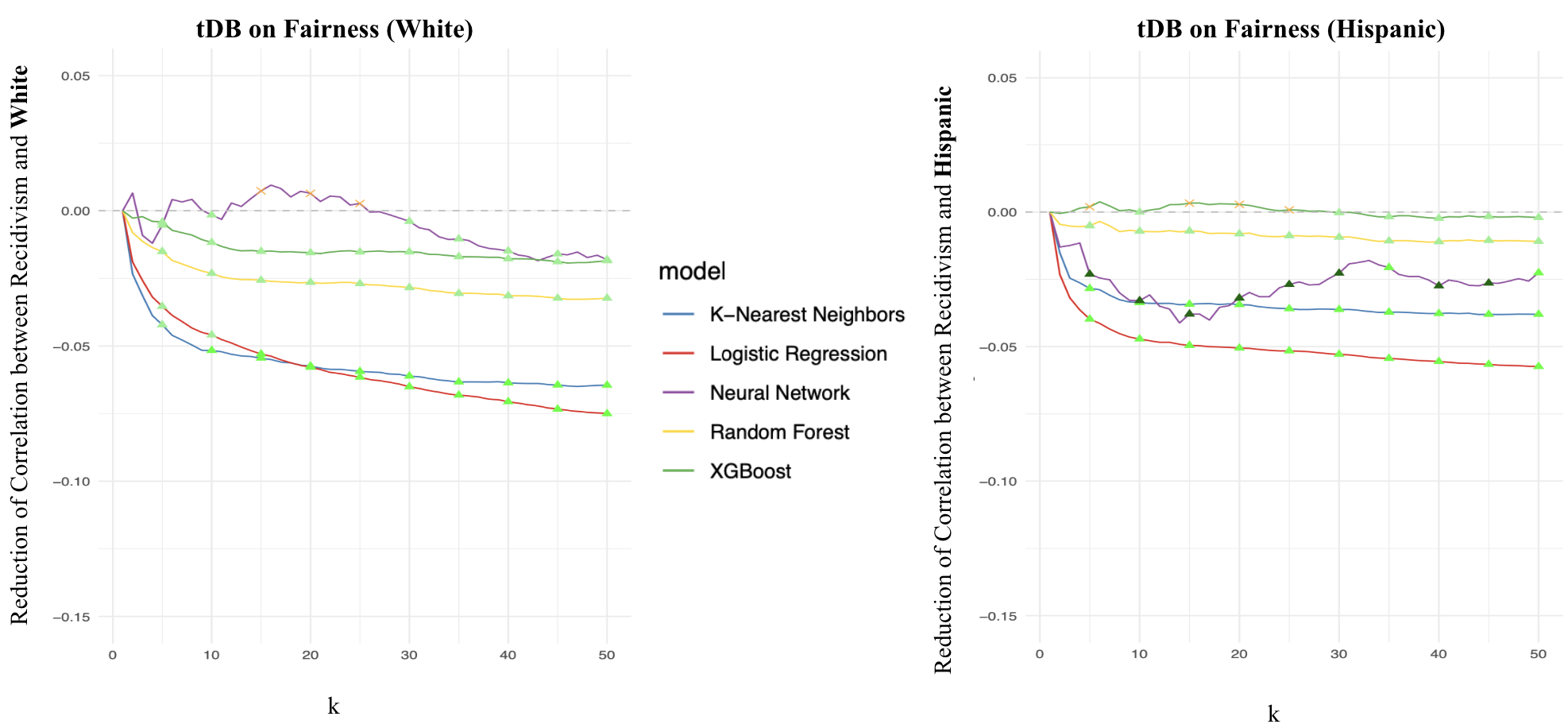}
    \vspace{0.5em}
    \caption{Effect of \emph{towerDebias} on Misclassification rate increase on \textbf{recidivism} and the correlation reductions with \textbf{race} in the COMPAS dataset.}
    \label{fig:cmp_ml} 
\end{figure}

\vspace{-1em}

Similar fairness patterns emerge across racial groups, with varying baseline correlations. 
Applying tDB to traditional machine learning models consistently reduces the correlation 
between predicted recidivism probability and each racial group by 25–50\%. For instance, 
the neural network model initially showed correlations of 0.24, 0.05, and 0.22 for Black, 
Hispanic, and White groups, respectively, with tDB achieving significant reductions. In 
terms of accuracy, the misclassification rate increases with higher values of $k$, with 
comparative accuracy losses being more pronounced than in regression tasks. 



\subsection{Iranian Churn}

\begin{figure}[H]
    \centering
    \caption*{ML versus \emph{towerDebias}: IranianChurn Results}
    \vspace{0.5em}
    \includegraphics[width=1\linewidth]{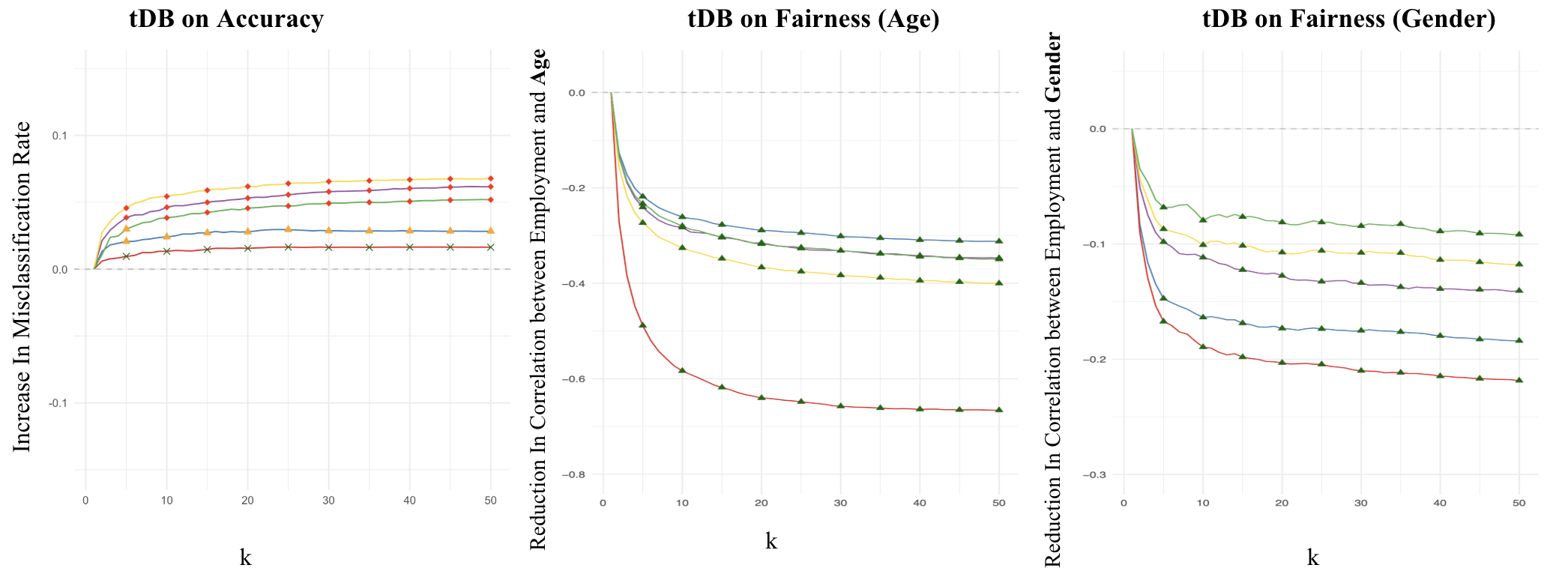}
    \caption{Effect of \emph{towerDebias} on Misclassification rate increase on \textbf{customer churn} and the correlation reductions with \textbf{gender \& age} in the IranChurn dataset.}
    \label{fig:iran_ml} 
\end{figure}

The \emph{Iranian Churn} dataset predicts customer churn, with ``Exited” (Yes/No) as 
the binary response variable and ``Gender” and ``Age” as sensitive variables. Each 
record contains features such as Credit Score, Geography, Gender, Age, Tenure, Balance, 
and Estimated Salary. Gender is binary, and Age is continuous. This setup allows us to 
further extend our analysis to a continuous sensitive variable beyond the categorical/binary 
case.

Results in Figure (\ref{fig:iran_ml}) show fairness improvements for both age and gender, 
with minimal accuracy losses. Consistent trends were observed for gender with previous datasets, 
and we see significant reductions in the correlation for age.  For example, the initial 
correlation between exit probability and age was 0.7 for the Logistic Regression model and 
around 0.35 for other methods. tDB consistently reduced this correlation to below 0.1 
starting at $k = 5$.

\subsection{Dutch Census}

The \emph{Dutch Census} dataset, collected by the Dutch Central Bureau
for Statistics in 2001, predicts whether an individual holds a
prestigious occupation (binary: Yes/No), with ``gender'' (binary:
male/female) as the sensitive attribute. The dataset includes 10 categorical predictors, 
which tDB encodes into 61 binary columns using one-hot encoding. This makes it a 
useful example for assessing tDB's effectiveness in sparse datasets. 

\begin{figure}[H]
    \centering    

    \caption*{ML versus \emph{towerDebias}: DutchCensus Results}
    \vspace{0.5em}
    \includegraphics[width=1\linewidth]{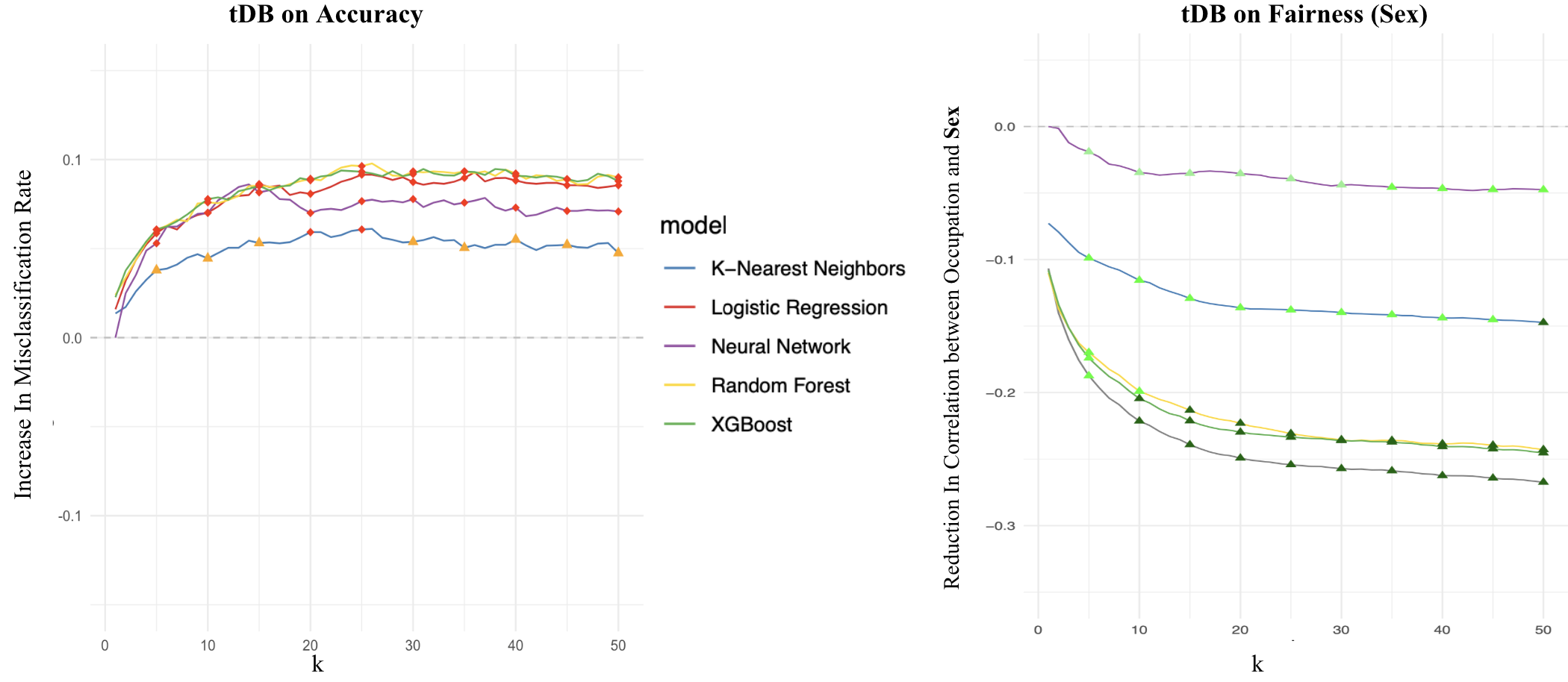}
    \vspace{-0.5em}
    \caption{Effect of \emph{towerDebias} on Misclassification rate increase on \textbf{employment status} and correlation reduction with \textbf{sex} in the DutchCensus dataset.}
    \label{fig:dutch_ml} 
\end{figure}
\vspace{-2em}

Results from figure (\ref{fig:dutch_ml}) are consistent with previous datasets, 
demonstrating fairness gains with some accuracy loss. Notably, correlation reductions 
of up to 50\% compared to baseline results highlight significant fairness improvements, 
though with considerably greater accuracy trade-offs for this specific dataset. 

Overall, the results from the \emph{ML versus towerDebias} plots 
indicate significant fairness improvements, though with some
potential accuracy losses in both regression and classification.

\subsection{FairML vs TowerDebias}

The fact that tDB is a post-processing approach to achieving fairness
suggests the possibility that it may be useful in improving the
performance of other fair ML methods. Here we present fairness-utility
results comparing tDB with FRRM and FGRRM algorithms.  The unfairness
parameter is set to (0.5, 0.2, 0.15, 0.1, and 0.05) to regulate baseline
fairness levels. Our goal is to assess how tDB further improves these
fairness results.

\vspace*{\fill}
\begin{figure}[H]
    \centering
    \includegraphics[width=1\linewidth]{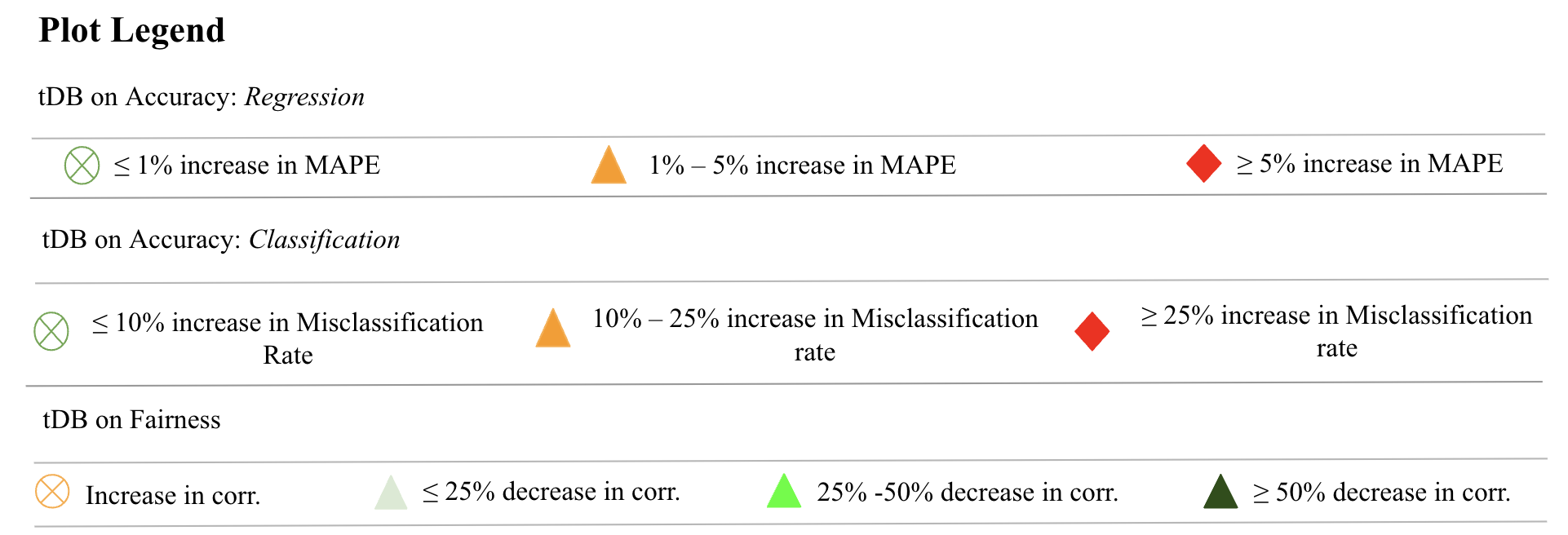}
\end{figure}

\subsubsection*{Regression}
\vspace{1em}

\begin{figure}[H]
    \centering
    \caption*{FairML versus \emph{towerDebias}: SVCensus Results}
    \vspace{0.5em}
    \includegraphics[width=1\linewidth]{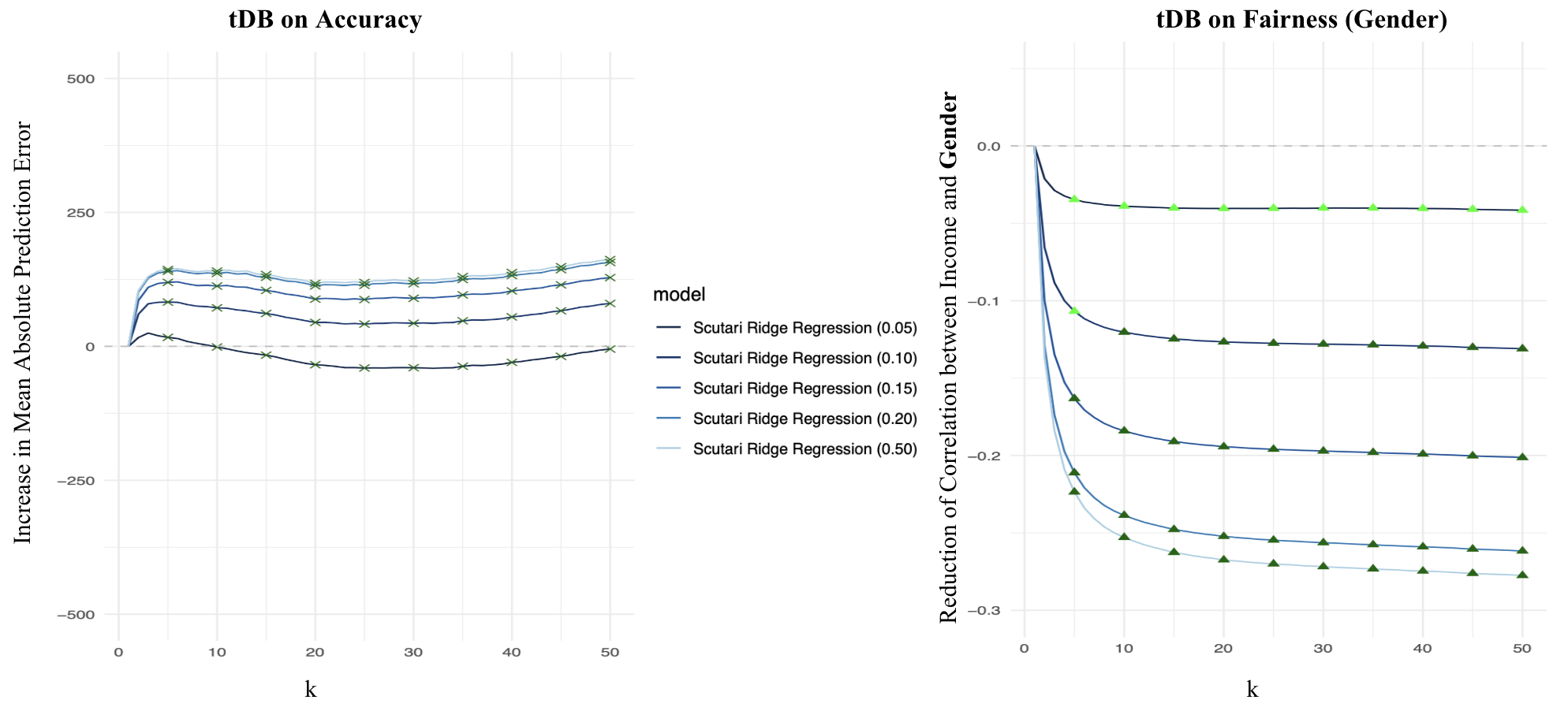}
    \vspace{-0.5em}
    \caption{Effect of \emph{towerDebias} on MAPE increase on predicting \textbf{wage income} and corresponding correlation reduction with \textbf{gender} in the SVCensus dataset.}
    \label{fig:svc_fairML} 
\end{figure}
\vspace*{\fill}

\newpage
\begin{figure}[H]
    \centering
    \caption*{FairML versus \emph{towerDebias}: Law School Admission Results}
    \vspace{0.5em}
    \includegraphics[width=0.95\linewidth]{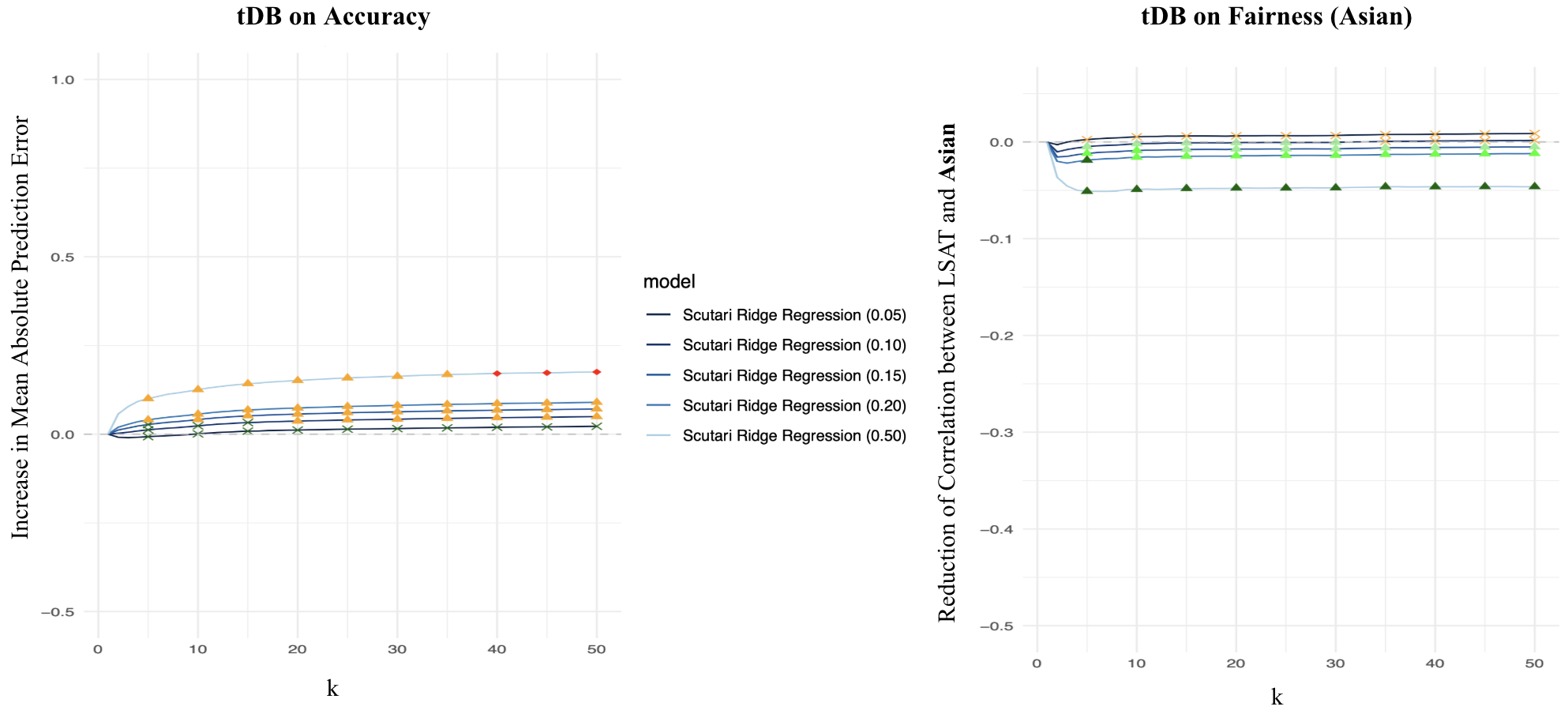}
    \vspace{-0.5em}
    \includegraphics[width=.95\linewidth]{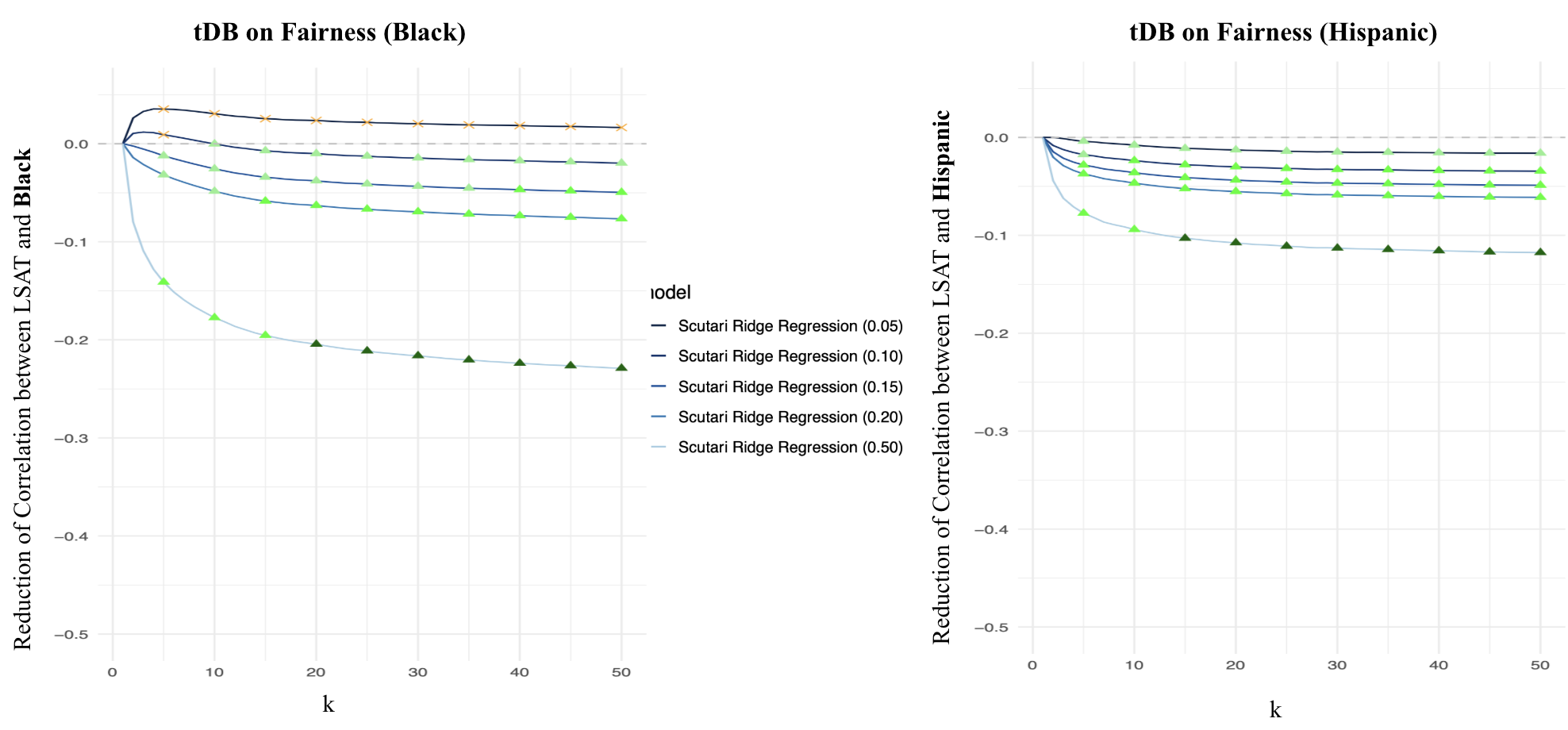}
    \vspace{-0.5em}
    \includegraphics[width=0.95\linewidth]{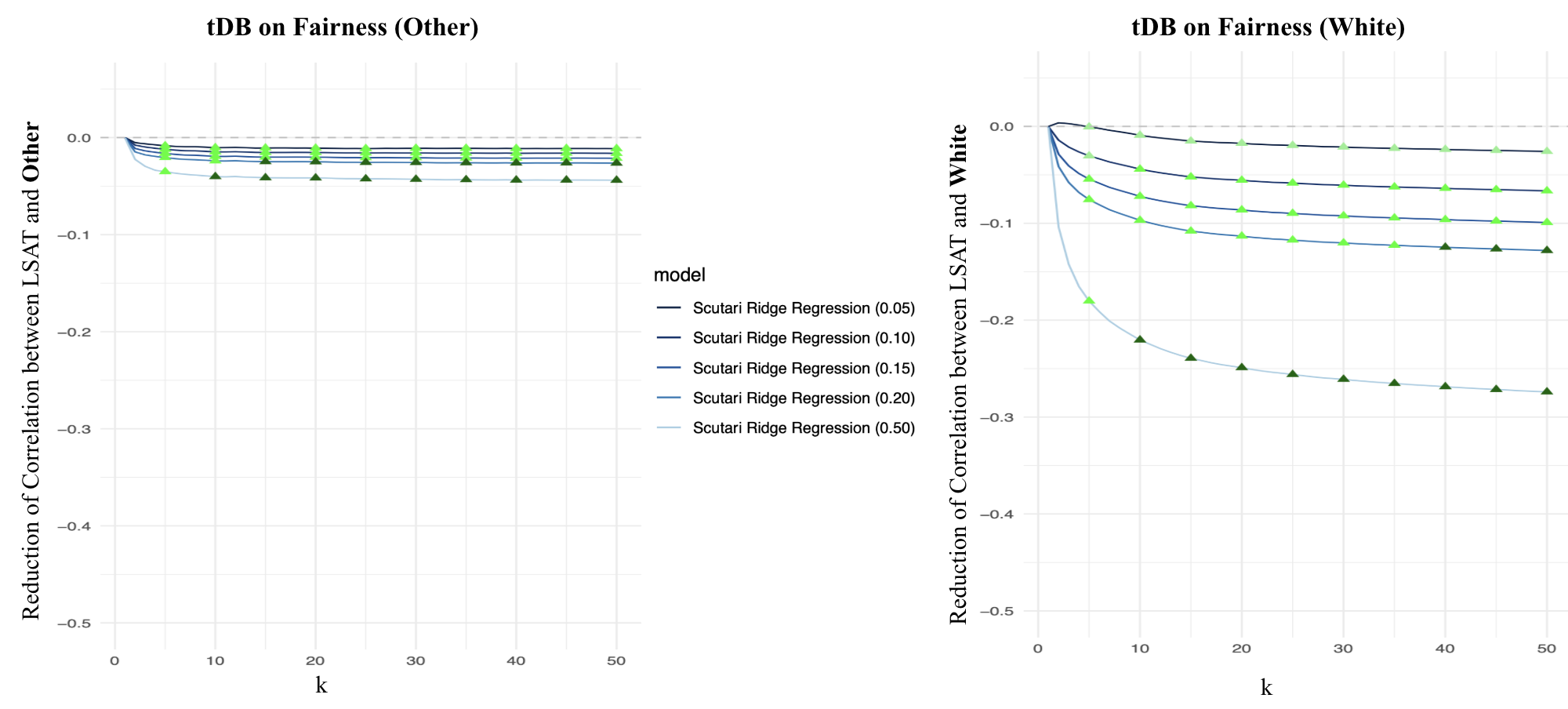}
    \vspace{0.5em}
    \caption{Effect of \emph{towerDebias} on MAPE increase on predicting \textbf{LSAT Score} and correlation reduction with \textbf{race} in the LSA dataset.}
    \label{fig:lsa_fairML} 
\end{figure}
\newpage

\vspace*{\fill}
\subsubsection*{Classification} 
\vspace{1em}

\begin{figure}[H]
    \centering
    \caption*{FairML versus \emph{towerDebias}: COMPAS Results}
    \vspace{0.5em}
    \includegraphics[width=1\linewidth]{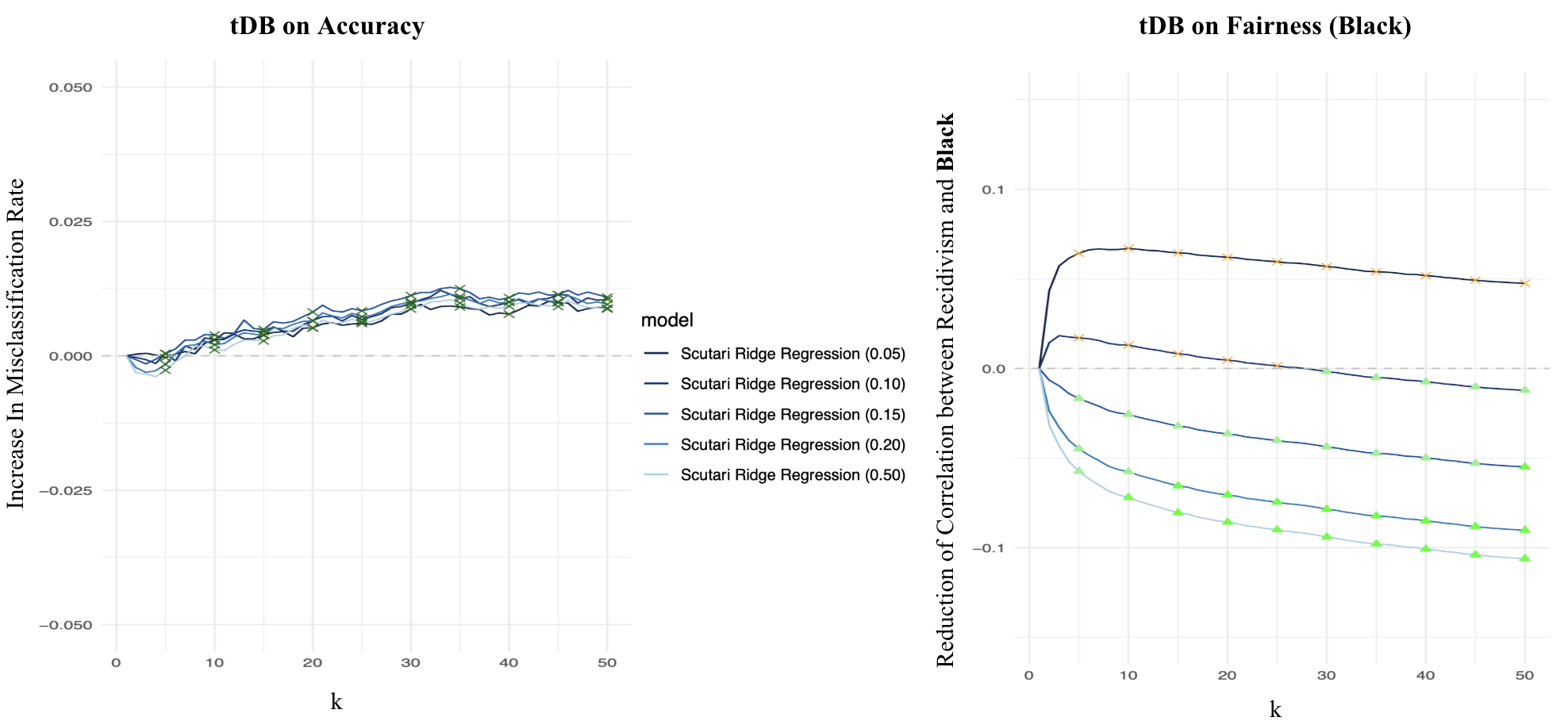}
    \vspace{2em}
    \vskip 0.1cm
    \includegraphics[width=1\linewidth]{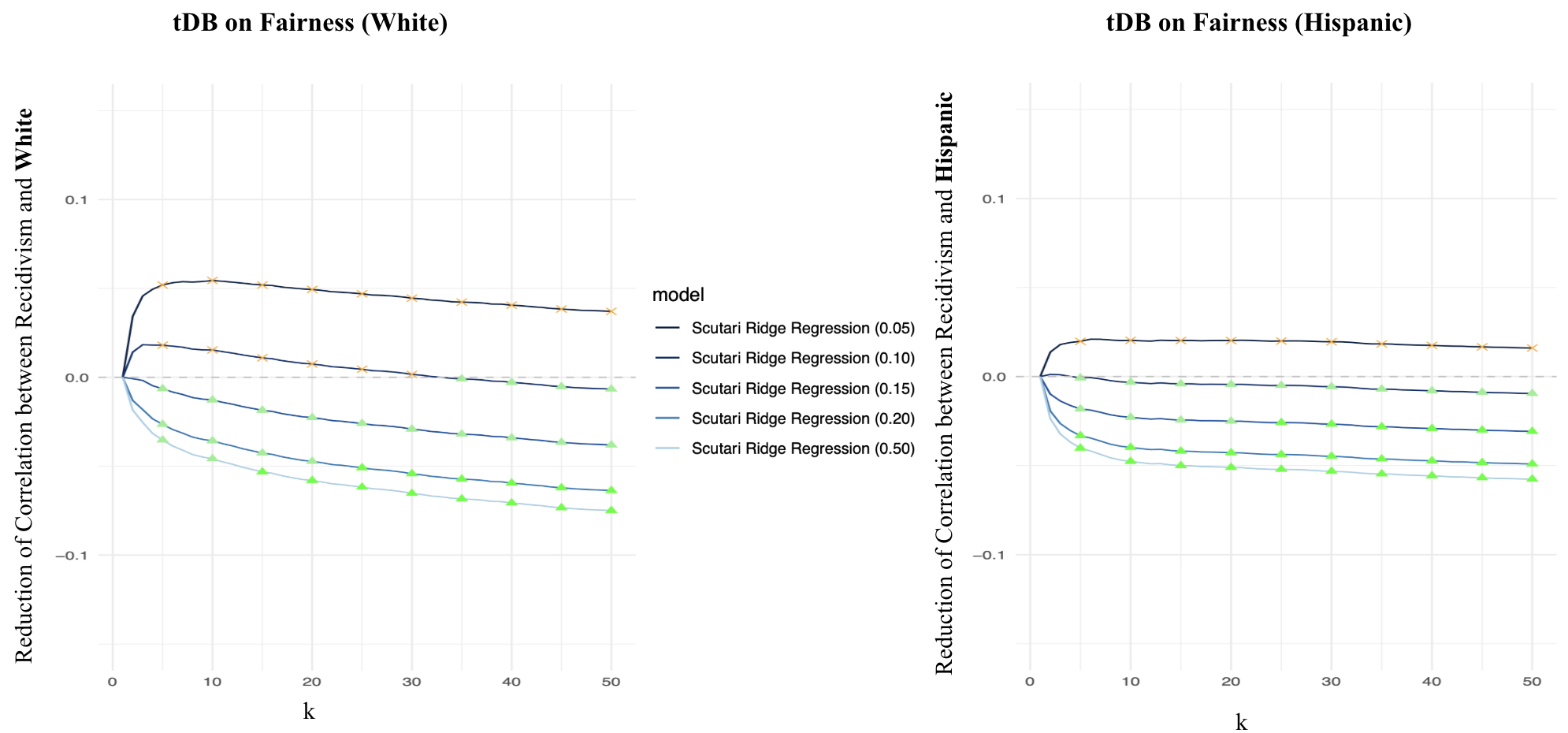}
    \vspace{0.25em}
    \caption{Effect of \emph{towerDebias} on Misclassification rate increase on \textbf{recidivism} and correlation reduction with \textbf{race} in the COMPAS dataset.}
    \label{fig:cmp_fairML} 
\end{figure}
\vspace*{\fill}

\newpage
\vspace*{\fill}
\begin{figure}[H]
    \centering
    \caption*{FairML versus \emph{towerDebias}: Iranian Churn Results}
    \vspace{0.5em}
    \includegraphics[width=1\linewidth]{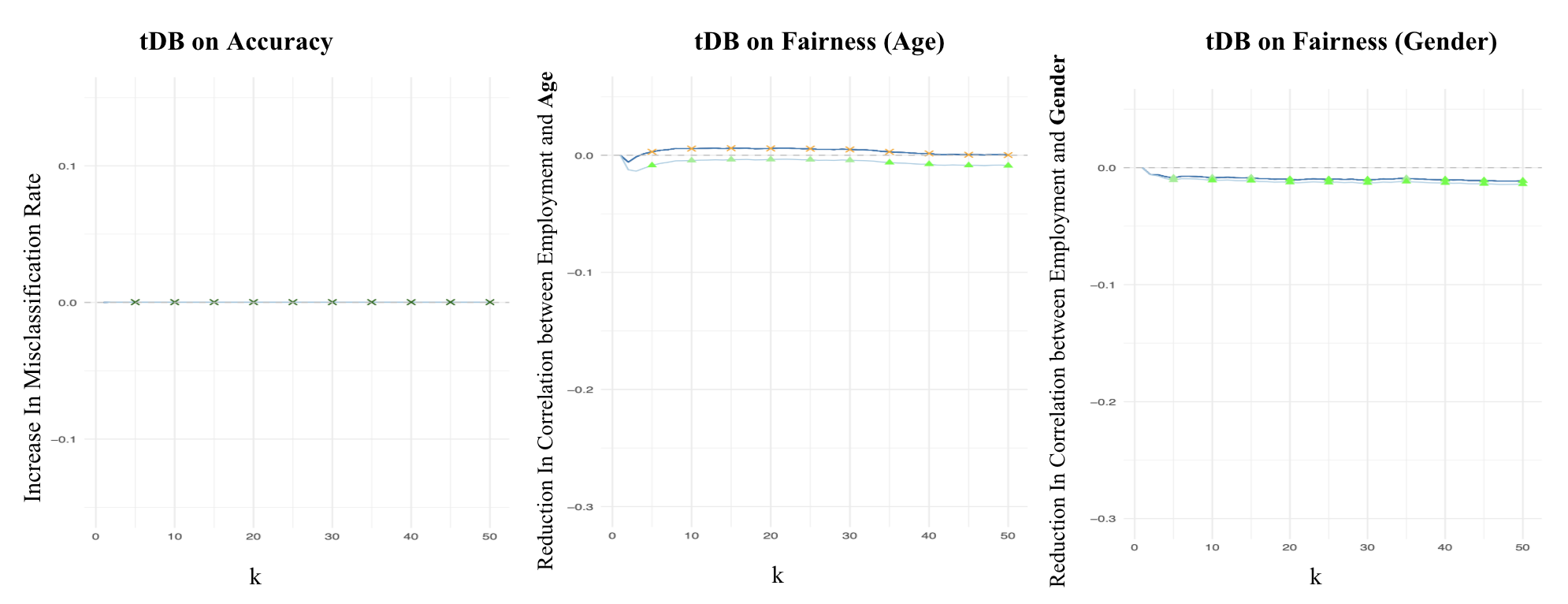}
    \caption{Effect of \emph{towerDebias} on Misclassification rate increase on \textbf{customer churn} and correlation reduction with \textbf{gender \& age} in the IranChurn dataset.}
    \label{fig:iran_fairML} 
\end{figure}

\vspace*{\fill}

\begin{figure}[H]
    \centering

    \caption*{FairML versus \emph{towerDebias}: DutchCensus Results}
    \vspace{0.5em}
    \includegraphics[width=1\linewidth]{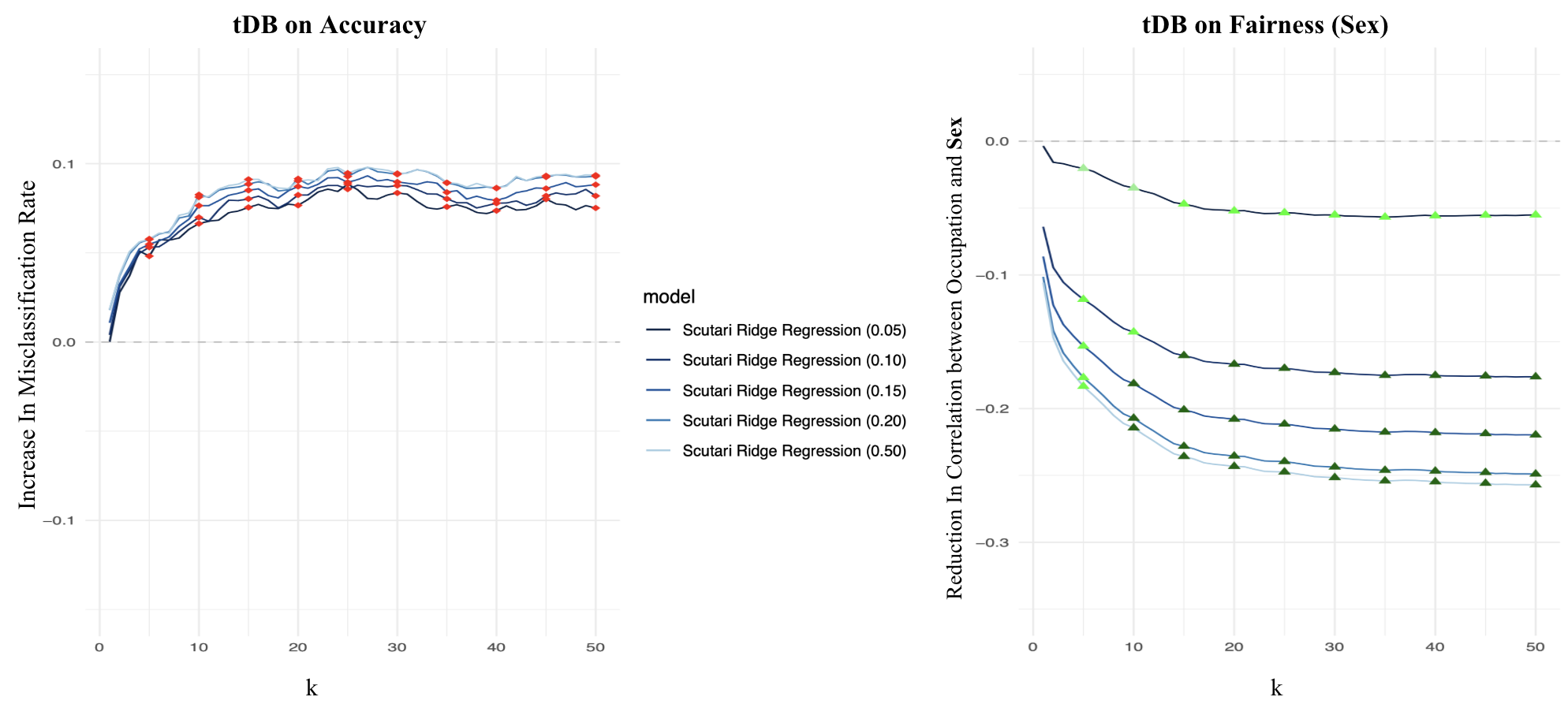}
    \vspace{-0.5em}
    \caption{Effect of \emph{towerDebias} on Misclassification rate increase on \textbf{employment status} and correlation reduction with \textbf{sex} in the DutchCensus dataset.}
    \label{fig:dutch_fairml} 
\end{figure}
\vspace*{\fill}
\newpage

From the \emph{FairML vs. towerDebias} plots, we observe that tDB
generally reduces the correlation between predictions and sensitive
variables across all datasets. As the unfairness parameter decreases (to
prioritize fairness), the initial correlation approach zero and are
considerably lower than the traditional models, with tDB further
improving these results. Fairness improvements are most pronounced (up
to 50\% reduction in correlation) for the ``less fair" baseline models,
although improvements are seen across all levels of unfairness. However,
for the COMPAS dataset in Figure (\ref{fig:cmp_fairML}), correlation
increases when using the most fair models (0.05/0.1), though overall,
consistent improvements are observed.

In some cases, \texttt{FairML} functions show considerably lower baseline
accuracy.  For example, the MAPE for FRRM functions in Figure
(\ref{fig:svc_fairML}) is \$35,000---over \$10,000 higher than
traditional ML algorithms. Further prioritizing fairness results in
additional accuracy loss.\footnote{Users should carefully balance
fairness and utility based on their specific needs.} The initial
correlation between predicted income and gender is minimized (0.08 at an
unfairness level of 0.05), but tDB continues to show further reductions
in correlation with minimal additional accuracy loss. Consistent trends are
observed across other datasets.

\subsubsection*{Choosing an appropriate $k$} 

The key question now is how to select an appropriate value of $k$ based
on the given application and the black-box method being used.

Our empirical results from the ML vs. tDB (and \texttt{FairML} vs tDB) plots
show that the choice of $k$ has considerable influence over
bias-variance trade-off, and consequently the fairness-utility
trade-off as well. For most applications, the results suggest that correlation
reductions appear to plateau after a certain point, usually starting at
low to moderate $k$ values. 

Let us consider the \emph{SVCensus} dataset. Starting $k$ = 7 or $k$ = 8
($k$ $\leq$ 10), the correlation reductions between predicted income and
gender seem to plateau at around 0.08-0.1, depending on the initial
algorithm. However, in terms of accuracy, we observe a consistent
increase in the mean absolute prediction error as $k$ increases. We may choose an early-to-moderate value of $k$ to maximize
the correlation reductions while minimizing corresponding accuracy
losses. Similar analyses can be conducted with other datasets.

\vspace{-1em}

\section{Discussion}\label{discussion}

In this paper, we introduce a novel post-processing approach to improve
prediction fairness of black-box machine learning algorithms. We propose 
\emph{towerDebias (tDB)}, a method that modifies black-box model predictions 
to estimate $E(Y|X)$ from an algorithm originally designed to predict 
$E(Y|X,S)$ to improve fairness while minimizing accuracy loss. Additionally,
we formally prove that applying this method reduces correlation and compare 
our fairness improvements to those in \cite{scutari2023fairml}. Our empirical 
evaluations of tDB across a variety of machine learning algorithms, datasets,
and $k$ values show promising results, with significant correlation 
reductions and minimal utility loss. These findings highlight the importance 
of selecting an appropriate $k$ to balance fairness improvements and 
predictive accuracy.

The importance of algorithmic fairness is clear across various applications,
as discussed by the COMPAS algorithm as an example of a black-box system with 
potential bias. Beyond COMPAS, tDB can be applied to a wide range of 
scenarios. Another example is Amazon’s biased hiring algorithm (2018) 
\citep{dastin2018amazon}, in which their AI-powered hiring tool exhibited 
bias against female candidates. Even if Amazon did not explicitly use 
“gender” as a feature, the presence of proxies—such as “women’s chess club”—
could indirectly reveal gender information. In such cases, the set $S$ in tDB
could be defined to include these proxies.\footnote{Regarding the issue of 
unbalanced data, this can be very complex, depending on factors such as the 
sampling methodology and definition of the target population. The issue, while 
certainly very important, is beyond the scope of this paper.} While 
domain-specific expertise is required to address more nuanced details,
tDB offers an efficient method for adjusting predictions with minimal computational 
or training costs. Its flexibility and ease of implementation make it a promising 
solution for mitigating algorithmic biases and enhancing fairness across a 
wide range of machine learning applications.

\section{Author Contributions}

The original idea was conceived by Dr. Norm Matloff. Aditya Mittal, an undergraduate Statistics major, 
made novel contributions to the graphical displays, conducted experiments, organized the source code, 
and authored the paper under Dr. Matloff's guidance. 

\bibliography{references}

\begin{thebibliography}{40}
\providecommand{\natexlab}[1]{#1}
\providecommand{\url}[1]{\texttt{#1}}
\expandafter\ifx\csname urlstyle\endcsname\relax
  \providecommand{\doi}[1]{doi: #1}\else
  \providecommand{\doi}{doi: \begingroup \urlstyle{rm}\Url}\fi

\bibitem[Agarwal et~al.(2019)Agarwal, Dud{\'\i}k, and Wu]{agarwal2019fair}
Alekh Agarwal, Miroslav Dud{\'\i}k, and Zhiwei~Steven Wu.
\newblock Fair regression: Quantitative definitions and reduction-based algorithms.
\newblock In \emph{International Conference on Machine Learning}, pages 120--129. PMLR, 2019.

\bibitem[Angwin and Larson(2016)]{propublica2}
Julia Angwin and Jeff Larson.
\newblock Machine bias: Technical response to northpointe, 2016.
\newblock URL \url{https://www.propublica.org/article/technical-response-to-northpointe}.

\bibitem[Angwin et~al.(2016)Angwin, Larson, Mattu, and Kirchner]{propublica1}
Julia Angwin, Jeff Larson, Surya Mattu, and Lauren Kirchner.
\newblock Machine bias, 2016.
\newblock URL \url{https://www.propublica.org/article/machine-bias-risk-assessments-in-criminal-sentencing}.

\bibitem[Axler(2015)]{axler}
Sheldon Axler.
\newblock \emph{Linear Algebra Done Right}.
\newblock Springer, 2015.

\bibitem[Baharlouei et~al.(2019)Baharlouei, Nouiehed, Beirami, and Razaviyayn]{baharlouei2019r}
Sina Baharlouei, Maher Nouiehed, Ahmad Beirami, and Meisam Razaviyayn.
\newblock R$\backslash$'enyi fair inference.
\newblock \emph{arXiv preprint arXiv:1906.12005}, 2019.

\bibitem[Barocas et~al.(2023)Barocas, Hardt, and Narayanan]{barocas-hardt-narayanan}
Solon Barocas, Moritz Hardt, and Arvind Narayanan.
\newblock \emph{Fairness and Machine Learning: Limitations and Opportunities}.
\newblock MIT Press, 2023.

\bibitem[Bourtoule et~al.(2021)Bourtoule, Chandrasekaran, Choquette-Choo, Jia, Travers, Zhang, Lie, and Papernot]{bourtoule2021machine}
Lucas Bourtoule, Varun Chandrasekaran, Christopher~A Choquette-Choo, Hengrui Jia, Adelin Travers, Baiwu Zhang, David Lie, and Nicolas Papernot.
\newblock Machine unlearning.
\newblock In \emph{2021 IEEE Symposium on Security and Privacy (SP)}, pages 141--159. IEEE, 2021.

\bibitem[Calmon et~al.(2017)Calmon, Wei, Vinzamuri, Natesan~Ramamurthy, and Varshney]{calmon2017optimized}
Flavio Calmon, Dennis Wei, Bhanukiran Vinzamuri, Karthikeyan Natesan~Ramamurthy, and Kush~R Varshney.
\newblock Optimized pre-processing for discrimination prevention.
\newblock \emph{Advances in neural information processing systems}, 30, 2017.

\bibitem[Chen et~al.(2024)Chen, Rossi, Park, Trivedi, Wang, Yu, Kim, Dernoncourt, and Ahmed]{chen2024fairness}
April Chen, Ryan~A Rossi, Namyong Park, Puja Trivedi, Yu~Wang, Tong Yu, Sungchul Kim, Franck Dernoncourt, and Nesreen~K Ahmed.
\newblock Fairness-aware graph neural networks: A survey.
\newblock \emph{ACM Transactions on Knowledge Discovery from Data}, 18\penalty0 (6):\penalty0 1--23, 2024.

\bibitem[Chen(2023)]{Zhisheng2023intro}
Zhisheng Chen.
\newblock Ethics and discrimination in artificial intelligence-enabled recruitment practices.
\newblock \emph{Humanities and Social Sciences Communications}, 10, 09 2023.
\newblock \doi{10.1057/s41599-023-02079-x}.

\bibitem[Chouldechova(2017)]{chouldechova2017fair}
Alexandra Chouldechova.
\newblock Fair prediction with disparate impact: A study of bias in recidivism prediction instruments.
\newblock \emph{Big data}, 5\penalty0 (2):\penalty0 153--163, 2017.

\bibitem[Cohen et~al.(2003)Cohen, Cohen, West, and Aiken]{cohen_applied_2003}
Jacob Cohen, Patricia Cohen, Stephen West, and Leona Aiken.
\newblock \emph{Applied Multiple Regression/Correlation Analysis for the Behavioral Sciences}.
\newblock Lawrence Erlbaum Associates, Mahwah, NJ, 3rd edition, 2003.
\newblock ISBN 978-0203774441.

\bibitem[Dastin(2018)]{dastin2018amazon}
Jeffrey Dastin.
\newblock Amazon scraps secret ai recruiting tool that showed bias against women.
\newblock \emph{Reuters}, October 2018.

\bibitem[Deho et~al.(2022)Deho, Zhan, Li, Liu, Liu, and le]{Deho2022intro}
Oscar Deho, Chen Zhan, Jiuyong Li, Jixue Liu, Lin Liu, and Thuc le.
\newblock How do the existing fairness metrics and unfairness mitigation algorithms contribute to ethical learning analytics?
\newblock \emph{British Journal of Educational Technology}, 53:\penalty0 1--22, 04 2022.
\newblock \doi{10.1111/bjet.13217}.

\bibitem[Denis et~al.(2024)Denis, Elie, Hebiri, and Hu]{JMLR:v25:23-0322}
Christophe Denis, Romuald Elie, Mohamed Hebiri, and Fran{\c{c}}ois Hu.
\newblock Fairness guarantees in multi-class classification with demographic parity.
\newblock \emph{Journal of Machine Learning Research}, 25\penalty0 (130):\penalty0 1--46, 2024.
\newblock URL \url{http://jmlr.org/papers/v25/23-0322.html}.

\bibitem[Durrett(2019)]{durrett}
R.~Durrett.
\newblock \emph{Probability: Theory and Examples}.
\newblock Cambridge Series in Statistical and Probabilistic Mathematics. Cambridge University Press, 2019.
\newblock ISBN 9781108473682.
\newblock URL \url{https://books.google.com/books?id=b22MDwAAQBAJ}.

\bibitem[Dwork et~al.(2012)Dwork, Hardt, Pitassi, Reingold, and Zemel]{dwork2012fairness}
Cynthia Dwork, Moritz Hardt, Toniann Pitassi, Omer Reingold, and Richard Zemel.
\newblock Fairness through awareness.
\newblock In \emph{Proceedings of the 3rd innovations in theoretical computer science conference}, pages 214--226, 2012.

\bibitem[Hardt et~al.(2016)Hardt, Price, and Srebro]{hardt2016equality}
Moritz Hardt, Eric Price, and Nati Srebro.
\newblock Equality of opportunity in supervised learning.
\newblock \emph{Advances in neural information processing systems}, 2016.
\newblock URL \url{https://arxiv.org/abs/1610.02413}.

\bibitem[Johndrow and Lum(2019)]{johndrow2019algorithm}
James~E Johndrow and Kristian Lum.
\newblock An algorithm for removing sensitive information.
\newblock \emph{The Annals of Applied Statistics}, 13\penalty0 (1):\penalty0 189--220, 2019.

\bibitem[Johnson and Wichern(1993)]{johnson1993mutli}
Richard Johnson and Dean Wichern.
\newblock \emph{Applied Multivariate Statistical Analysis}.
\newblock Pearson, 1993.

\bibitem[Karimi et~al.(2022)Karimi, Akbar~Khan, Liu, Derr, and Liu]{10032333}
Hamid Karimi, Muhammad~Fawad Akbar~Khan, Haochen Liu, Tyler Derr, and Hui Liu.
\newblock Enhancing individual fairness through propensity score matching.
\newblock In \emph{2022 IEEE 9th International Conference on Data Science and Advanced Analytics (DSAA)}, pages 1--10, 2022.
\newblock \doi{10.1109/DSAA54385.2022.10032333}.

\bibitem[Komiyama et~al.(2018)Komiyama, Takeda, Honda, and Shimao]{pmlr-v80-komiyama18a}
Junpei Komiyama, Akiko Takeda, Junya Honda, and Hajime Shimao.
\newblock Nonconvex optimization for regression with fairness constraints.
\newblock In Jennifer Dy and Andreas Krause, editors, \emph{Proceedings of the 35th International Conference on Machine Learning}, volume~80 of \emph{Proceedings of Machine Learning Research}, pages 2737--2746. PMLR, 10--15 Jul 2018.
\newblock URL \url{https://proceedings.mlr.press/v80/komiyama18a.html}.

\bibitem[Kozodoi et~al.(2022)Kozodoi, Jacob, and Lessmann]{kozodoi2022fairness}
Nikita Kozodoi, Johannes Jacob, and Stefan Lessmann.
\newblock Fairness in credit scoring: Assessment, implementation and profit implications.
\newblock \emph{European Journal of Operational Research}, 2022.

\bibitem[Lee et~al.(2022)Lee, Bu, Sattigeri, Panda, Wornell, Karlinsky, and Feris]{lee2022maximal}
Joshua Lee, Yuheng Bu, Prasanna Sattigeri, Rameswar Panda, Gregory Wornell, Leonid Karlinsky, and Rogerio Feris.
\newblock A maximal correlation approach to imposing fairness in machine learning.
\newblock In \emph{ICASSP 2022-2022 IEEE International Conference on Acoustics, Speech and Signal Processing (ICASSP)}, pages 3523--3527. IEEE, 2022.

\bibitem[Lum and Johndrow(2016)]{lum2016statistical}
Kristian Lum and James Johndrow.
\newblock A statistical framework for fair predictive algorithms.
\newblock \emph{arXiv preprint arXiv:1610.08077}, 2016.

\bibitem[Madras et~al.(2018)Madras, Creager, Pitassi, and Zemel]{madras2018learning}
David Madras, Elliot Creager, Toniann Pitassi, and Richard Zemel.
\newblock Learning adversarially fair and transferable representations.
\newblock In \emph{International Conference on Machine Learning}, pages 3384--3393. PMLR, 2018.

\bibitem[Mary et~al.(2019)Mary, Calauz{\`e}nes, and Karoui]{Mary2019FairnessAwareLF}
J{\'e}r{\'e}mie Mary, Cl{\'e}ment Calauz{\`e}nes, and Noureddine~El Karoui.
\newblock Fairness-aware learning for continuous attributes and treatments.
\newblock In \emph{International Conference on Machine Learning}, 2019.
\newblock URL \url{https://api.semanticscholar.org/CorpusID:174800179}.

\bibitem[Matloff and Zhang(2022)]{matloff2022novel}
Norman Matloff and Wenxi Zhang.
\newblock A novel regularization approach to fair ml.
\newblock \emph{arXiv preprint arXiv:2208.06557}, 2022.

\bibitem[Roh et~al.(2023)Roh, Lee, Whang, and Suh]{roh2023improving}
Yuji Roh, Kangwook Lee, Steven~Euijong Whang, and Changho Suh.
\newblock Improving fair training under correlation shifts.
\newblock In \emph{International Conference on Machine Learning}, pages 29179--29209. PMLR, 2023.

\bibitem[Sarker(2021)]{Sarker2019intro}
Iqbal Sarker.
\newblock Machine learning: Algorithms, real-world applications and research directions.
\newblock 2021.
\newblock URL \url{https://pubmed.ncbi.nlm.nih.gov/33778771/}.

\bibitem[Scutari(2023)]{scutari2023fairml}
Marco Scutari.
\newblock fairml: A statistician's take on fair machine learning modelling.
\newblock \emph{arXiv preprint arXiv:2305.02009}, 2023.
\newblock URL \url{https://arxiv.org/abs/2305.02009}.

\bibitem[Silvia et~al.(2020)Silvia, Ray, Tom, Aldo, Heinrich, and John]{Silvia2020post}
Chiappa Silvia, Jiang Ray, Stepleton Tom, Pacchiano Aldo, Jiang Heinrich, and Aslanides John.
\newblock A general approach to fairness with optimal transport.
\newblock \emph{Proceedings of the AAAI Conference on Artificial Intelligence}, 34:\penalty0 3633--3640, 04 2020.
\newblock \doi{10.1609/aaai.v34i04.5771}.

\bibitem[Wehner and Köchling(2020)]{Wehner2020intro}
Marius Wehner and Alina Köchling.
\newblock Discriminated by an algorithm: A systematic review of discrimination and fairness by algorithmic decision-making in the context of hr recruitment and hr development.
\newblock \emph{BuR - Business Research}, pages 1--54, 11 2020.
\newblock \doi{10.1007/s40685-020-00134-w}.

\bibitem[Wi{\'s}niewski and Biecek(2021)]{wisniewski2021fairmodels}
Jakub Wi{\'s}niewski and Przemys{\l}aw Biecek.
\newblock fairmodels: A flexible tool for bias detection, visualization, and mitigation.
\newblock \emph{arXiv preprint arXiv:2104.00507}, 2021.

\bibitem[Wolpert(2009)]{wolpert2009tower}
Robert Wolpert.
\newblock Institute of statistics and decision sciences, 2009.
\newblock URL \url{https://www2.stat.duke.edu/courses/Spring09/sta205/lec/topics/rn.pdf}.

\bibitem[Zafar et~al.(2019)Zafar, Valera, Gomez-Rodriguez, and Gummadi]{JMLR:v20:18-262}
Muhammad~Bilal Zafar, Isabel Valera, Manuel Gomez-Rodriguez, and Krishna~P. Gummadi.
\newblock Fairness constraints: A flexible approach for fair classification.
\newblock \emph{Journal of Machine Learning Research}, 20\penalty0 (75):\penalty0 1--42, 2019.
\newblock URL \url{http://jmlr.org/papers/v20/18-262.html}.

\bibitem[Zemel et~al.(2013)Zemel, Wu, Swersky, Pitassi, and Dwork]{pmlr-v28-zemel13}
Rich Zemel, Yu~Wu, Kevin Swersky, Toni Pitassi, and Cynthia Dwork.
\newblock Learning fair representations.
\newblock In Sanjoy Dasgupta and David McAllester, editors, \emph{Proceedings of the 30th International Conference on Machine Learning}, volume~28 of \emph{Proceedings of Machine Learning Research}, pages 325--333, Atlanta, Georgia, USA, 17--19 Jun 2013. PMLR.
\newblock URL \url{https://proceedings.mlr.press/v28/zemel13.html}.

\bibitem[Zhang and Ntoutsi(2019)]{zhang2019faht}
Wenbin Zhang and Eirini Ntoutsi.
\newblock Faht: an adaptive fairness-aware decision tree classifier.
\newblock \emph{arXiv preprint arXiv:1907.07237}, 2019.

\bibitem[Zhang et~al.(2021)Zhang, Bifet, Zhang, Weiss, and Nejdl]{zhang2021farf}
Wenbin Zhang, Albert Bifet, Xiangliang Zhang, Jeremy~C Weiss, and Wolfgang Nejdl.
\newblock Farf: A fair and adaptive random forests classifier.
\newblock In \emph{Pacific-Asia conference on knowledge discovery and data mining}, pages 245--256. Springer, 2021.

\bibitem[Zhao et~al.(2022)Zhao, Dai, Shu, and Wang]{zhao2022towards}
Tianxiang Zhao, Enyan Dai, Kai Shu, and Suhang Wang.
\newblock Towards fair classifiers without sensitive attributes: Exploring biases in related features.
\newblock In \emph{Proceedings of the Fifteenth ACM International Conference on Web Search and Data Mining}, pages 1433--1442, 2022.

\end{thebibliography}

\end{document}